%% file: main.tex
\pdfoutput=1

\documentclass{article}
\usepackage[utf8]{inputenc}

\PassOptionsToPackage{numbers, compress}{natbib}
 \usepackage[main, final]{neurips_2026}

\input{preamble/preamble.tex}

\title{SKIM: Pruning Large Language Model Agents via Selective Knowledge Informed Masking}
\author{
  Moonseok Choi\\
  KAIST \\
  \texttt{ms.choi@kaist.ac.kr}\\
  \And
  Jongwon Jeong\\
  University of Wisconsin–Madison\\
  \texttt{jjeong64@wisc.edu}\\
  \And
  Minki Kang \\
  KAIST \\
  \texttt{minkikang@kaist.ac.kr}
  \And
  Giung Nam$^\dagger$\\
  KAIST\\
  \texttt{giung@kaist.ac.kr}\\
  \And
  Juho Lee$^\dagger$\\
  KAIST\\
  \texttt{juholee@kaist.ac.kr} \\
}

\begin{document}
\maketitle
\begingroup
\renewcommand{\thefootnote}{\fnsymbol{footnote}}
\footnotetext[2]{Co-corresponding authors.}
\endgroup

\input{main/abstract}
\input{main/introduction}
\input{main/related_works}
\input{main/preliminaries}
\input{main/skim}
\input{main/experiments}
\input{main/discussion}
\clearpage
\newpage
\section*{Acknowledgements}
This work was supported by Institute of Information \& communications Technology Planning \& Evaluation(IITP) grant funded by the Korea government(MSIT) (No.RS-2019-II190075, Artificial Intelligence Graduate School Program(KAIST)). This work was supported by the National Research Foundation of Korea(NRF) grant funded by the Korea government(MSIT) (RS-2022-NR070855). This work was supported by the Institute of Information \& Communications Technology Planning \& Evaluation (IITP) with a grant funded by the Ministry of Science and ICT (MSIT) of the Republic of Korea in connection with the Global AI Frontier Lab International Collaborative Research Agreement (No. RS-2024-00469482 \& RS-2024-00509279). We declare no competing interests.

\clearpage
\newpage
\appendix
\clearpage
\newpage
\input{appendix/related_works}
\input{appendix/method_details}
\input{appendix/further_experiments}
\input{appendix/prompts}
\clearpage
\newpage
\input{appendix/checklist}

\end{document}

%% file: preamble/preamble.tex
\usepackage{wrapfig}
\usepackage[format=hang]{subcaption}
\usepackage{colortbl}
\usepackage[dvipsnames,table]{xcolor}
\usepackage{multirow}

\usepackage[T1]{fontenc}
\usepackage{times}
\usepackage{amsmath, amssymb}
\usepackage{graphicx}

\newcommand{\sk}{\cellcolor{blue!5}}

\usepackage{calc}
\newsavebox\CBox
\def\BF#1{\sbox\CBox{#1}\resizebox{\wd\CBox}{\ht\CBox}{\textbf{#1}}}
\def\UL#1{\underline{#1}}
\def\BL#1{\BF{\UL{#1}}}

\newcommand{\cell}[2]{%
  \makebox[\widthof{00.0}][r]{#1}%
  \,\textcolor{gray!90}{\scriptsize/\,%
  \makebox[\widthof{\scriptsize 00.0}][r]{#2}}%
}
\newcommand{\bcell}[2]{%
  \BL{\makebox[\widthof{00.0}][r]{#1}}%
  \,\textcolor{gray!90}{\scriptsize/\,%
  \makebox[\widthof{\scriptsize 00.0}][r]{#2}}%
}
\newcommand{\dcell}[2]{%
  \textit{\makebox[\widthof{00.0}][r]{#1}}%
  \,\textcolor{gray!90}{\scriptsize/\,%
  \makebox[\widthof{\scriptsize 00.0}][r]{\textit{#2}}}%
}
\definecolor{decreasecolor}{HTML}{B22222}
\definecolor{increasecolor}{HTML}{1F4E79}

\usepackage{tcolorbox}
\tcbuselibrary{skins}
\definecolor{promptbg}{HTML}{F5F5F5}
\definecolor{promptfr}{HTML}{9E9E9E}

\usepackage{booktabs, array}
\usepackage{listings}

\definecolor{citecolor}{RGB}{0,102,204}
\definecolor{linkcolor}{RGB}{190,105,30}
\definecolor{urlcolor}{RGB}{199,21,133}

\usepackage[colorlinks,linktoc=all]{hyperref}
\usepackage[all]{hypcap}
\hypersetup{citecolor=citecolor}
\hypersetup{linkcolor=linkcolor}
\hypersetup{urlcolor=urlcolor}
\usepackage[nameinlink,capitalise]{cleveref}
\creflabelformat{equation}{#2\textup{#1}#3}
\crefname{section}{\S}{\S\S}

\lstdefinestyle{mystyle}{
    commentstyle=\color{OliveGreen},
    numberstyle=\tiny\color{black!60},
    stringstyle=\color{BrickRed},
    basicstyle=\ttfamily\scriptsize,
    breakatwhitespace=false,
    breaklines=true,
    captionpos=b,
    keepspaces=true,
    numbers=none,
    numbersep=5pt,
    showspaces=false,
    showstringspaces=false,
    showtabs=false,
    tabsize=2
}

%% file: main/abstract.tex
\begin{abstract}
Large Language Model (LLM) agents are increasingly deployed under structured generation, where grammar constraints enforce output format at inference time.
Whereas existing compression methods target unconstrained generation, we first identify a fundamental mismatch: under structured generation, grammar constraints reduce the model's output distribution to valid tokens, and the logits matter most at the small set of positions that decide the agent's action.
Motivated by this finding, we propose SKIM (Selective Knowledge-Informed Masking), the first pruning and distillation framework for LLM agents under structured generation, with both stages guided by token role.
SKIM partitions token positions into context, decision, and format categories, and this partition drives two complementary axes: (1) category-aware pruning saliency scores that preserve parameters most relevant for decision positions; and (2) an on-policy distillation objective that aligns student and teacher on valid tokens, paired with category-aware feature matching.
We evaluate SKIM on tool-calling benchmarks and within multi-step agent loops, across multiple model scales and sparsity levels.
SKIM consistently improves accuracy over baselines, with its superiority becoming most striking in high-sparsity regimes where conventional methods struggle, demonstrating the potential of category-aware compression for deployment-time grammar-constrained agents.
The project page is available at \href{https://moon4sake.github.io/skim}{link}.

\end{abstract}

%% file: main/introduction.tex
\section{Introduction}
\label{sec:intro}

\begin{wrapfigure}[11]{r}{0.29\textwidth}
  \centering
  \vspace{-2.8em}
  \includegraphics[width=\linewidth]{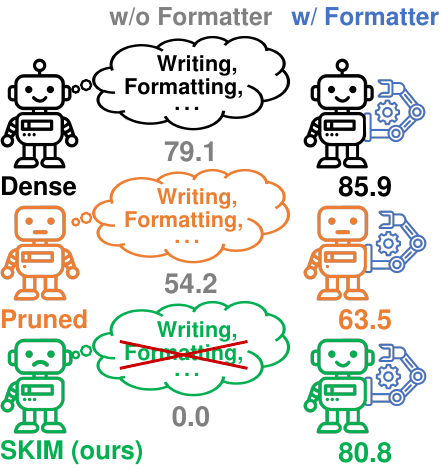}
  \caption{Target behavior.}
  \label{fig:skima}
\end{wrapfigure}

Large Language Model (LLM) agents are widely served under structured generation, where outputs must conform to a predefined format and tokens that would violate it are masked before sampling at every decoding step~\citep{willard2023efficient,dong2025xgrammar}.
This is now standard practice in production serving stacks~\citep{dong2025xgrammar,zheng2024sglang}.
Yet the pruning and distillation methods used to compress LLMs are designed under the assumption of unconstrained generation~\citep{frantar2023sparsegpt,sun2023simple,gu2024minillm,muralidharan2024compact}, and recent work confirms that off-the-shelf compression degrades sharply on tool-calling tasks~\citep{dong2025can}.

The mismatch is \emph{structural}, not incidental.
Under structured generation, the model's effective distribution at each step is its raw softmax restricted to valid tokens and renormalized~\citep{willard2023efficient,geng2025jsonschemabench}.
The objectives used in prior compression work, KL divergence to a teacher~\citep{gu2024minillm,muralidharan2024compact} and layer-wise activation reconstruction~\citep{frantar2023sparsegpt,sun2023simple}, are all computed against the unconstrained output.
They preserve parameters that contribute to outputs the runtime masks, wasting capacity on behavior it already enforces.

\begin{figure}[t]
    \centering
    \includegraphics[width=\linewidth]{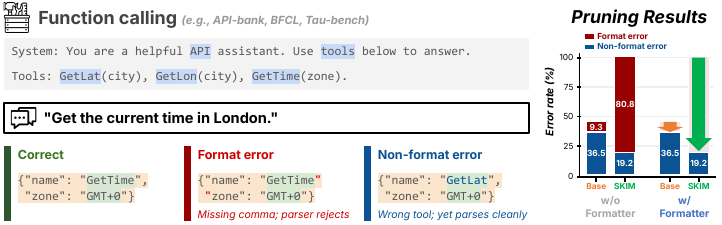}
    \caption{
    Token categories and error decomposition under structured
    generation. \textbf{Left:} Example function call with tokens highlighted by category ({\setlength{\fboxsep}{0pt}\colorbox[HTML]{C9DAF8}{\emph{Context}}}, {\setlength{\fboxsep}{0pt}\colorbox[HTML]{D9EAD3}{\emph{Decision}}}, {\setlength{\fboxsep}{0pt}\colorbox[HTML]{FCE5CD}{\emph{Format}}}) and three response types distinguishing format errors (eliminated by structured decoding) from non-format errors (which require the model itself to be correct). \textbf{Right:} Pruned-then-distilled model performance on Qwen3.5-2B~\citep{qwen3.5} and BFCL-v3~\citep{patil2025berkeley} at 25\% sparsity. SKIM fully exploits the grammar formatter: without it, SKIM looks broken because format errors dominate, but with it, SKIM easily beats Base (Wanda-sp). Format errors are essentially free under structured decoding, so reducing non-format errors at decision positions is what matters.
    }
    \label{fig:intro}
    \vspace{5pt}
\end{figure}

Our key insight is that \emph{not all output tokens play the same role under structured generation.}
Hinging on this insight, we partition the agent's output into three roles: \emph{(1) context} positions hold the input the model reads, \emph{(2) decision} positions are the few output points where the format admits multiple valid options such as the function name or enum-valued arguments, and \emph{(3) format} positions are the syntactic scaffolding the constraint mechanism enforces~\citep{dong2025xgrammar,geng2025jsonschemabench}.
The constraint mechanism enforces format validity for free at deployment~\citep{willard2023efficient,dong2025xgrammar}, so capacity allocated to format positions is wasted on behavior the runtime supplies regardless.
\Cref{fig:intro}~(right) shows that even with standard distillation for healing~\citep{muralidharan2024compact,sreenivas2024llm}, the error of pruned agent models is dominated by non-format error, indicating that compression should preserve capacity at decision rather than format positions.

\setlength{\textfloatsep}{17pt}

We propose \textbf{SKIM (Selective Knowledge-Informed Masking)}, the first pruning and distillation framework designed for LLM agents deployed under structured generation.
The pruning side extends layer-wise reconstruction with per-category activation coefficients that
down-weight format positions in the saliency
score~\citep{sun2023simple,an2024fluctuation}.
The distillation side combines a forward-KL objective between the teacher's and student's constrained distributions on the student's own on-policy rollouts, paired with hidden-state matching at context and decision positions~\citep{muralidharan2024compact}.
Both stages share the same token-role partition, aligning the preserved capacity with the deployment regime end-to-end.

We evaluate SKIM on tool-calling benchmarks, BFCL-v3~\citep{patil2025berkeley} and API-Bank~L1~\citep{li2023api}, and within ReAct-style agent loops~\citep{yao2022react} on HotpotQA~\citep{yang2018hotpotqa}.
Across model scales from 0.8B to 9B and sparsity levels, SKIM generally outperforms strong pruning and distillation baselines~\citep{sun2023simple,an2024fluctuation,ling2024slimgpt}, with the largest margins at high sparsity where existing methods collapse but SKIM continues to recover most of the dense model's structured-decoding accuracy.

Our contributions are:
\begin{itemize}
    \setlength\itemsep{0pt}
    \item To the best of our knowledge, we are the first to identify the mismatch between standard pruning and distillation objectives and constrained-decoding deployment, and we introduce a token-wise categorization of agent outputs into context, decision, and format positions, which grounds the design of our method (\cref{sec:prelim}).
    \item We propose SKIM, a pruning and distillation framework that uses this categorization in both stages: a category-aware saliency that down-weights format positions in layer-wise reconstruction, and an on-policy distillation objective over the teacher's and student's constrained distributions paired with category-aware hidden-state matching (\cref{sec:skim}).
    \item We evaluate SKIM on tool-calling benchmarks and within multi-step agent loops across model scales from 0.8B to 9B, showing that SKIM preserves reasoning and tool-use capability with gains widening at higher sparsities (\cref{sec:experiments}).
\end{itemize}

%% file: main/related_works.tex
\section{Related Work}
\label{sec:related}

\paragraph{Tool-calling LLM agents.}
Large language models are increasingly deployed as autonomous agents interacting with external APIs \citep{yao2022react, schick2023toolformer, patil2024gorilla, qin2023toolllm, zhang2025xlam, li2023api}. As these agents transition to production-scale deployments, they face intertwined bottlenecks in efficiency and reliability. While prior work improved robustness via tool embeddings \citep{hao2023toolkengpt} or function masking \citep{lin2024hammer}, reliable deployment fundamentally requires syntactically flawless API requests (e.g., strict JSON). Because even minor structural deviations cause cascading execution failures, production pipelines overwhelmingly rely on mechanisms that explicitly enforce the desired output structure.

\paragraph{Structured generation.}
Structured generation enforces a grammar at inference time, guaranteeing structurally valid outputs.
Early constrained-decoding algorithms~\citep{deutsch2019general,geng2023grammar} formalized this idea, and modern engines~\citep{beurer2023prompting, willard2023efficient, lew2023sequential} realize it efficiently by masking invalid tokens against predefined schemas or finite-state machines, which we call a \emph{formatter} in \cref{fig:intro,fig:skima}.
Their objective is faithful, low-overhead grammar enforcement, and the LLM is treated as a black box that the engine wraps without modification.
A grammar, however, makes some token positions deterministic and others not - at deterministic positions the constraint uniquely fixes the next token, while at the rest the model's prediction selects the output.
This position-level asymmetry is outside the scope of grammar-enforcement engines, but for compression methods designed for the structured-generation regime it is exactly the signal worth exploiting for compression.

\paragraph{LLM pruning.}
LLM pruning methods divide along granularity and saliency framework. Unstructured pruning methods score individual weights for sparse
patterns~\citep{frantar2023sparsegpt,sun2023simple}, while structured pruning methods remove whole architectural units for wall-clock speedups on dense hardware~\citep{ma2023llm,ashkboos2024slicegpt,men2025shortgpt,an2024fluctuation,ling2024slimgpt}.
Cutting across both is \emph{layer-wise reconstruction}, an influential saliency framework that scores units by how much zeroing them perturbs the layer's output on calibration activations~\citep{frantar2023sparsegpt,sun2023simple,an2024fluctuation,ling2024slimgpt}.
SKIM extends layer-wise reconstruction by partitioning calibration
activations by the role each position plays under structured
generation.
We accordingly select Wanda~\citep{sun2023simple}, FLAP~\citep{an2024fluctuation}, and SlimGPT~\citep{ling2024slimgpt}
as baselines, spanning unstructured and structured layer-wise
reconstruction methods and providing the most direct comparison to
SKIM.

\paragraph{LLM distillation.}
Knowledge distillation (KD) facilitates knowledge transfer from a teacher to a student model by matching output distributions or intermediate hidden states~\citep{hinton2015distilling,romero2014fitnets}.
While KD has long been a well-established paradigm in language modeling~\citep{jiao2020tinybert,sanh2019distilbert,sun2019patient}, recent research has shifted toward on-policy approaches, where students learn from their own generated trajectories to improve alignment in generative tasks~\citep{gu2024minillm,agarwal2024policy,yang2025qwen3}.
Additionally, KD has been utilized to recover performance degradation following LLM pruning~\citep{muralidharan2024compact,sreenivas2024llm}.
However, to the best of our knowledge, no existing studies have explored KD specifically within the context of grammar-constrained decoding.
For the detailed literature review, we refer readers to \cref{app:further_related_works}.

%% file: main/preliminaries.tex
\section{Preliminaries and Motivation}
\label{sec:prelim}

\subsection{Preliminaries}
\label{sec:prelim-background}

\paragraph{Constrained language generation.}
Modern autoregressive language models define the conditional probability of a sequence $y \equiv y_{1:T} = (y_{1},\dots,y_{T})$ given an initial prompt $x$ as:
\begin{align}\textstyle
p(y_{1:T} \mid x) = \prod_{t=1}^{T} p(y_{t} \mid x, y_{<t}),
\end{align}
where $x$ and $y$ are sequences composed of tokens from a vocabulary $\mathcal{V}$, and generation proceeds from left to right.
To incorporate specific requirements into the generation process (e.g., grammar constraints in formal language), we assume a potential function $\phi$ that assigns scores to sequences based on desired attributes.
To satisfy constraints during generation, a widely adopted technique is \emph{locally constrained decoding}, which steers the generation process at each time step $t$ by drawing tokens from a re-weighted density:
\begin{align}\textstyle
p_{\phi}(y_{t} \mid x, y_{<t}) \propto p(y_{t} \mid x, y_{<t}) \frac{\phi(x,y_{1:t})}{\phi(x,y_{<t})}.
\label{eq:constrained-dist}
\end{align}
In this work, we focus on \emph{binary hard constraints} where $\phi \in \{0, 1\}$.
We assume $\phi$ is monotonic such that if a prefix violates the constraints, any of its extensions will also be invalid, i.e., $\phi(x, y_{<t}) = 0$ implies $\phi(x,y_{1:t}) = 0$ for all $y_{t} \in \mathcal{V}$.
Under this assumption, the ratio $\phi(x, y_{1:t}) / \phi(x, y_{<t})$ is safely defined as zero whenever the denominator is zero.
Such a condition is typically satisfied in \emph{structured generation} tasks, where syntactic or formal constraints can be verified incrementally.

\textit{Remark.}
In practice, structured generation libraries~\citep{willard2023efficient,dong2025xgrammar} enforce these constraints by defining a valid token set $\mathcal{V}_{t} \equiv \mathcal{V}(x,y_{<t}) = \{ y_{t} \in \mathcal{V} \mid \phi(x,y_{1:t}) = 1\}$ at each time step $t$.
Tokens not in $\mathcal{V}_{t}$ are masked by setting their logits to $-\infty$, restricting the sampling space of $p(\cdot \mid x, y_{<t})$ to $\mathcal{V}_{t}$.

\paragraph{Token position categories.}
For a sequence $(x, y)$ comprising prompt $x$ and output $y$, we partition token positions into three functional categories, \emph{context} ($c$), \emph{decision} ($d$), and \emph{format} ($f$), as illustrated in \cref{fig:intro}.
\emph{Context positions} $P_{c}(x)$ are prompt-side positions, specifically the tokens of tool names, descriptions, and user queries.
In contrast, \emph{decision positions} $P_{d}(y) = \{t \in [T] : |\mathcal{V}_t| > 1\}$ are output positions where the schema leaves more than one valid continuation, requiring the model to choose among alternatives, whereas \emph{format positions} $P_{f}(y) = \{t \in [T] : |\mathcal{V}_t| = 1\}$ are output positions where the schema admits a unique continuation, rendering the conditional distribution $p_\phi(\cdot \mid x, y_{<t})$ a point mass independent of the model's logits.

\subsection{Motivation: format-position capacity is recoverable under structured decoding}
\label{sec:motivation}

Let $(x, y^{\ast})$ be a prompt and its corresponding reference response pair, where $\phi(x, y^{\ast})=1$.
The cross-entropy of the $\phi$-constrained neural network model $p_{\phi}$ with respect to this pair is given by:
\begin{align}\textstyle\label{eq:structured_cross_entropy_loss}
\mathcal{L}_{\text{CE}}(p_{\phi} ; x, y^{\ast})
= -\sum_{t=1}^{T} \log{p_{\phi}(y_{t}^{\ast} \mid x, y_{<t}^{\ast})},
\end{align}
which is the quantity that must be preserved across both pruned and distilled models.
Notably, for any format position $t \in P_{f}(y^{\ast})$, the prediction is deterministic, such that $p_{\phi}(y_{t} \mid x, y_{<t}^{\ast}) = [y_{t} = y_{t}^{\ast}]$ for all $y_{t} \in \mathcal{V}$.
Consequently, both the gradient and the Hessian of the loss with respect to any internal weight matrix $W$ evaluate to zero at these specific positions:
\begin{align}
\forall t \in P_{f}(y^{\ast}) :
\nabla_{W}[-\log{p_{\phi}(y_{t}^{\ast} \mid x, y_{<t}^{\ast})}] = 0
\;\;\text{and}\;\;
\nabla^{2}_{W}[-\log{p_{\phi}(y_{t}^{\ast} \mid x, y_{<t}^{\ast})}] = 0.
\end{align}
While this property is absent in the layer-wise reconstruction loss typically used as a proxy in standard LLM pruning~\citep{frantar2023sparsegpt,sun2023simple}, it reveals a fundamental asymmetry across token positions when evaluated under the structured cross-entropy loss in \cref{eq:structured_cross_entropy_loss}.
Hinging on this, we investigate whether such positional asymmetry can be leveraged even within standard LLM pruning scenarios.

\paragraph{Empirical results.}
In addition to \cref{fig:intro} (right), \cref{app:fig:error_decomposition} shows the error decomposition aggregated over BFCL and API-Bank, two Qwen3.5 model sizes (0.8B and 2B), and three pruning baselines (Wanda~\citep{sun2023simple}, FLAP~\citep{an2024fluctuation}, SlimGPT~\citep{ling2024slimgpt}).
Each baseline applies its own pruning mask and then heals the pruned model via knowledge distillation on xLAM-function-calling data~\citep{liu2024apigen} under a matched compute budget; structured generation at inference uses XGrammar~\citep{dong2025xgrammar}.
Total error grows with sparsity, but non-format error dominates at every level: format errors are largely absorbed by the constraint mechanism, while non-format errors (wrong-tool-or-argument failures that determine the agent's action) remain the principal source of error.
In \cref{app:motivating_experiment}, we further corroborate this pattern through a direct ablation that
zeroes units specialized for each token category: removing
format-specialized capacity barely changes accuracy, while removing
context- or decision-specialized capacity collapses it
(see \cref{app:fig:motivation_ablation}).
Together, these results suggest substantial room for improvement at decision-relevant positions, motivating SKIM's category-aware design.

\paragraph{Evaluation protocols.}
Throughout the paper, we adopt two metrics based on full-match accuracy against reference tool-call sequences for evaluation:
(1) \emph{structured accuracy} (i.e., $y \sim p_{\phi}$), which reflects the deployment setting under schema-constrained decoding and measures practical performance;
(2) \emph{free-generation accuracy} (i.e., $y \sim p$), a diagnostic under unconstrained decoding that isolates the model's intrinsic capabilities from the gains provided by the constraint mechanism $\phi$.

\textit{Remark.}
A narrow gap between these metrics suggests that the model has natively dedicated its capacity to satisfying the constraints.
Conversely, a large gap suggests it relies on $\phi$ to enforce the format, decoupling its intrinsic reasoning from the required structure.

%% file: main/skim.tex
\section{SKIM}
\label{sec:skim}

\subsection{Layer-wise pruning with position-awareness}
\label{sec:skim-pruning}

\paragraph{Reconstruction loss.}
Let $X \in \mathbb{R}^{T \times M}$ be the input activations and $W \in \mathbb{R}^{M \times N}$ be the original dense weight matrix in a neural network layer.
The standard \emph{layer-wise reconstruction loss} is:
\begin{align}\label{eq:layerwise_reconstruction_loss}
\mathcal{L}(\hat{W} ; W, X) = \left\lVert XW - X\hat{W} \right\rVert_{F}^{2},
\end{align}
which measures how well the pruned weights $\hat{W}$ preserve the layer outputs.
If we view the loss as a function of $\smash{\mathrm{vec}(\hat{W})}$, its full Hessian is given by $\nabla^{2} \mathcal{L} = I_{N} \otimes H$ where $H:=2X^{\top}X$.
Under the diagonal approximation $H \approx \mathrm{diag}(H)$, we ignore cross-feature interactions and obtain:
\begin{align}\textstyle\label{eq:reconstruction_loss_surrogate}
\mathcal{L}(\hat{W} ; W, X) \approx \sum_{i=1}^{M} \mathcal{L}_{i},
\quad\text{where }\;
\mathcal{L}_{i} = \frac{1}{2}H_{ii} \left\lVert W_{i,:} - \hat{W}_{i,:} \right\rVert_{2}^{2}
\text{ and }\;
H_{ii} = 2\sum_{t=1}^{T}X_{ti}^{2}.
\end{align}

\paragraph{Gaussian interpretation.}
From a probabilistic perspective, \cref{eq:reconstruction_loss_surrogate} admits a Gaussian interpretation; up to an additive constant, $\mathcal{L}_{i}$ is the negative log-likelihood of the isotropic Gaussian model:
\begin{align}
W_{i,:} \mid \hat{W}_{i,:}, X_{:,i} \sim \mathcal{N}(\hat{W}_{i,:}, \sigma_{i}^{2} I_{N}),
\end{align}
with precision $1/\sigma_{i}^{2} = H_{ii}$.
Indeed, $\mathcal{L}_{i}$ corresponds to the quadratic term of the log-likelihood:
\begin{align}\textstyle
\log p(W_{i,:} \mid \hat{W}_{i,:}, X_{:,i}) =
- \frac{N}{2}\log(2 \pi \sigma_{i}^{2})
- \frac{1}{2\sigma_{i}^{2}} \left\lVert W_{i,:} - \hat{W}_{i,:} \right\rVert_{2}^{2},
\end{align}
which implies that a larger sum of squared activations assigns a smaller variance, making the likelihood more sensitive to perturbations in the corresponding row of the weight matrix.
The saliency of the $i^{\text{th}}$ neuron is then evaluated as the decrease in the log-likelihood when it is zeroed ($\smash{\hat{W}_{i,:}=0}$), which exactly corresponds to the square of the heuristic pruning metric proposed in \citep{sun2023simple}:
\begin{align}\textstyle\label{eq:standard_saliency}
S_{i}
= \frac{1}{2} H_{ii} \left\lVert W_{i,:} \right\rVert_{2}^{2}
= \left( \sum_{t=1}^{T} X_{ti}^{2} \right) \left\lVert W_{i,:} \right\rVert_{2}^{2}.
\end{align}

\paragraph{Position-aware saliency score.}
However, \cref{eq:standard_saliency} weights all token positions uniformly through the sum over $t$.
Motivated by the observation that token contributions along the sequence dimension are not uniform, we introduce a context-weighted heteroscedastic surrogate precision:
\begin{align}
\tilde{H}_{ii}^{\text{pos}}
:= 2 \cdot \textstyle\left(
    \alpha_{c} \sum_{t \in P_{c}} X_{t i}^{2}
    + \alpha_{d} \sum_{t \in P_{d}} X_{ti}^{2}
    + \alpha_{f} \sum_{t \in P_{f}} X_{t i}^{2}
\right),
\end{align}
where $P_{c}, P_{d}, P_{f}$ form exclusive positional groups corresponding to context, decision, and format token roles, respectively, and $\alpha_{c}, \alpha_{d}, \alpha_{f} \geq 0$.
Using $\smash{1/\sigma_{i}^{2} := \tilde{H}_{ii}^{\text{pos}}}$ in the saliency formulation yields the corresponding position-aware saliency score:
\begin{align}\textstyle\label{eq:context_aware_saliency_score}
S_{i}^{\text{pos}}
= \left(
    \alpha_{c} \sum_{t \in P_{c}} X_{t i}^{2}
    + \alpha_{d} \sum_{t \in P_{d}} X_{t i}^{2}
    + \alpha_{f} \sum_{t \in P_{f}} X_{t i}^{2}
\right)
\cdot
\left\lVert W_{i,:} \right\rVert_{2}^{2}.
\end{align}

\textit{Remark.}
Notably, \cref{eq:context_aware_saliency_score} can be interpreted as being derived from a weighted layer-wise reconstruction loss, i.e., $\smash{\mathcal{L}(\hat{W} ; W, X) := \lVert A^{1/2}(XW-X\hat{W}) \rVert_{F}^{2}}$, where $A$ is a diagonal scaling matrix with entries $A_{tt} = \alpha_{c}[t \in P_{c}] + \alpha_{d}[t \in P_{d}] + \alpha_{f}[t \in P_{f}]$.
Nevertheless, a key advantage of our formulation lies in the modularity of the surrogate precision.
By decoupling position-awareness from the loss objective and embedding it directly into the precision term, our framework is not restricted to any specific gating mechanism.
Instead, it establishes a generalized design space: one can readily define diverse functional forms for $\smash{\tilde{H}_{ii}^{\text{pos}}}$ to incorporate different positional inductive biases, enabling the systematic derivation of a wide family of position-aware saliency scores without altering the underlying standard layer-wise reconstruction loss in \cref{eq:layerwise_reconstruction_loss}.
In what follows, we experimentally compare various surrogate precision designs through an ablation study in \cref{app:pruning-details}.

\subsection{SKIM Distillation}
\label{sec:skim-kd}

\paragraph{Distillation loss.}
Let $p$ be the original dense model and $q$ be the pruned sparse model.
In practice, to recover the performance degradation from pruning, knowledge distillation~\citep{hinton2015distilling} is often employed by minimizing a forward KL divergence between the two models on a dataset of $(x, y)$ pairs~\citep{muralidharan2024compact}:
\begin{align}\textstyle\label{eq:kd-vanilla}
    D_{\text{KL}}(p \parallel q)(y \mid x)
    = \sum_{t=1}^{T} D_{\text{KL}}(p(\cdot \mid x, y_{<t}) \parallel q(\cdot \mid x, y_{<t})).
\end{align}
In a constrained generation setup, however, the distillation loss should be reformulated to minimize the divergence between the $\phi$-constrained distributions, $p_{\phi}$ and $q_{\phi}$:
\begin{align}\textstyle
D_{\text{KL}}(p_{\phi} \parallel q_{\phi})(y \mid x)
= \sum_{t=1}^{T} D_{\text{KL}}(p_{\phi}(\cdot \mid x, y_{<t}) \parallel q_{\phi}(\cdot \mid x, y_{<t})),
\end{align}
where the $t^{\text{th}}$ term in the summation decomposes as
\begin{align}\label{eq:kd-decomposition}\textstyle
\frac{1}{Z_p} \sum_{y \in \mathcal{V}_t} p(y \mid x,y_{<t}) \left[ \log \frac{p(y \mid x,y_{<t})}{q(y \mid x,y_{<t})} - \log Z_p + \log Z_q \right],
\end{align}
where $Z_{p}$ and $Z_{q}$ are the normalization constants for $p_{\phi}$ and $q_{\phi}$, required to yield valid probability distributions from the unnormalized formulations in \cref{eq:constrained-dist}.
Notably, the logit masking technique for structured generation remains applicable here, allowing for the direct computation of the KL divergence over the valid vocabulary $\mathcal{V}_{t}$ without explicitly evaluating the normalization constants:
\begin{align}\textstyle\label{eq:kd-ours}
D_{\text{KL}}(p_{\phi} \parallel q_{\phi})(y \mid x)
= -\sum_{t=1}^{T} \sum_{y_{t} \in \mathcal{V}_{t}} p_{\phi}(y_{t} \mid x, y_{<t}) \log{q_{\phi}(y_{t} \mid x, y_{<t})} + \mathrm{const}.
\end{align}

\textit{Remark.}
The decomposition in Eq.~\ref{eq:kd-decomposition} clarifies why the constrained form is appropriate here.
While the gradient of the first term in the bracket with respect to the $k^{\text{th}}$ logit of $q$ is non-zero for all $k$, the cancellation of $\log{Z}$ terms yields zero gradients for the logits corresponding to $\mathcal{V} \setminus \mathcal{V}_{t}$.
Consequently, the loss only shapes the student's distribution over $\mathcal{V}_{t}$, which is the only factor that matters during inference under structured decoding.
Such surgical specificity is crucial for tool-calling LLMs where shared parameters govern both constrained and unconstrained generation.
Unlike vanilla distillation in \cref{eq:kd-vanilla}, which unnecessarily suppresses off-support tokens, our approach in \cref{eq:kd-ours} avoids such updates, preventing performance leakage into unconstrained behaviors like free-form chat.

While \cref{eq:kd-vanilla,eq:kd-ours} minimize forward KL divergence to align token-level distributions, they leave open various design choices.
Key among these are the source of $y$ (e.g., original dataset, teacher-, or student-generated) and the specific divergence measure (e.g., forward or reverse KL) employed for matching~\citep{agarwal2024policy}.
As $\mathcal{V}_t$ and the later decision positions depend on the student's earlier decisions, we adopt, after an off-policy warm-up, an \emph{on-policy forward KL} on student-generated responses $y$:
\begin{align}
\mathcal{L}_{\text{KD}}(q ; p)
= \mathbb{E}_{x \in \mathcal{D}} \mathbb{E}_{y \sim \texttt{sg}(q_{\phi}(\cdot \mid x))} \left[
    D_{\text{KL}}(p_{\phi} \parallel q_{\phi})(y \mid x)
\right].
\end{align}
We further adopt feature distillation~\citep{romero2014fitnets,lu2022knowledge,muralidharan2024compact}, but refine its scope to focus primarily on \emph{context} and \emph{decision} token positions.
Indeed, our empirical findings in \cref{sec:exp-ablation} indicate that down-weighting \emph{format} positions improves results, aligning with our argument that the limited capacity of the pruned model should be strategically allocated.
The final distillation loss for SKIM is formulated as:
\begin{equation}\label{eq:skim-loss}\textstyle
\mathcal{L}_{\text{SKIM}}
= \mathcal{L}_{\text{KD}} + \lambda_H \cdot \mathcal{L}_{\text{hidden}},
\quad\text{where }
\mathcal{L}_{\text{hidden}}
= \sum_{\ell \in L} \sum_{t} \alpha_{\mathrm{cat}(t)}
    \cdot \big\| h_t^{(p, \ell)} - h_t^{(q, \ell)} \big\|_2^2,
\end{equation}
where $h_t^{(p, \ell)}$ and $h_t^{(q, \ell)}$ are the teacher and student hidden states, respectively, at position $t$ following transformer block $\ell$.
For further empirical details, we refer readers to \cref{app:kd-details}.

%% file: main/experiments.tex
\section{Experiments}
\label{sec:experiments}

\subsection{Experimental setup}
\label{sec:exp-setup}

\paragraph{Models and datasets.}
We evaluate SKIM across four scales of the Qwen3.5 series~\citep{qwen3.5}: Qwen3.5-0.8B/2B/4B/9B.
Dense models serve as reference.
xLAM-function-calling-60k~\citep{liu2024apigen} provides activation statistics for pruning calibration and training data for distillation.
We evaluate on two tool-calling benchmarks: BFCL-v3~\citep{patil2025berkeley} and API-Bank L1~\citep{li2023api}, reporting full-match accuracy against reference tool calls.
For the agentic pipeline evaluation, we use HotpotQA~\citep{yang2018hotpotqa} dataset under ReAct-style loop~\citep{yao2022react}, reporting F1 metric. Numbers are reported under structured decoding (the deployment setting) and free generation (see \cref{sec:motivation}).
All decoding is greedy ($\tau = 0$).

\paragraph{Sparsity levels.}
We report sparsity at 10\%, 25\%, and 50\%, applied as structured sparsity over the prunable-unit set (MLP output channels, attention KV groups, and linear-attention groups).
Following the convention in prior structured-pruning work~\citep{an2024fluctuation, ling2024slimgpt}, these rates are
computed relative to the prunable parameters only; modules outside this set (token embeddings and the LM head) are not pruned.
The corresponding total parameter reductions are roughly 6--7\%, 15--18\%, and 30--36\% across the four model scales -- for instance, $6.77\%$, $17.07\%$, and $34.14\%$ on Qwen3.5-2B (see \cref{app:efficiency} for the full parameter and memory breakdown).

\paragraph{Baselines.}
Following \cref{sec:related}, we compare against three layer-wise reconstruction baselines that share SKIM's framework and provide the most direct comparison: Wanda-sp (the structured variant of Wanda~\citep{sun2023simple}), FLAP~\citep{an2024fluctuation}, and
SlimGPT~\citep{ling2024slimgpt}.
Each baseline is pruned with its own saliency criterion and then healed with standard output-matching knowledge distillation on the fixed xLAM-FC subset, under matched training compute.
As Qwen3.5 series are not tool-specialists, we fine-tune dense models as well for a fair comparison across methods.

\paragraph{SKIM settings.}
We use category coefficients $(\alpha_c, \alpha_d, \alpha_f) = (1.0, 1.0, 0.2)$, fixed across models and sparsity levels, and unnormalized feature-distillation weight $\lambda_H = 0.5$.
We label context token categories using GPT-4~\citep{openai2024gpt4technicalreport}.
Distillation is run for 2 epochs at a learning rate of $1 \times 10^{-5}$ on a 4{,}096-sample subset of xLAM-FC, with maximum sequence length 1{,}300 and maximum generation length 300.
Detailed hyperparameters and implementation specifics are deferred to \cref{app:implementation_details}.

\begin{table}[t]
\centering
\caption{Post-KD full-match accuracy of Qwen3.5 series across varying sparsity ratios with thinking mode enabled.
All reported numbers are measured under the same distillation budget.
Each cell reports \textbf{structured-generation} \textcolor{gray!90}{\scriptsize/\,free-generation} accuracy.
SKIM achieves the best structured-generation accuracy in most settings, with the gap to baselines widening at higher sparsity.
}
\label{tab:main}
\setlength{\aboverulesep}{0pt}
\setlength{\belowrulesep}{0pt}
\setlength{\extrarowheight}{0.15em}
\setlength{\tabcolsep}{0.5em}
\renewcommand{\arraystretch}{1.1}
\resizebox{\textwidth}{!}{\begin{tabular}{>{\centering\arraybackslash}p{0.7cm}p{3.2cm}*{6}{>{\centering\arraybackslash}p{1.6cm}}}
    \toprule
    & & \multicolumn{3}{c}{BFCL-v3} & \multicolumn{3}{c}{API-Bank L1} \\
    \cmidrule(lr){3-5} \cmidrule(lr){6-8}
    & Method~/~Sparsity & $10\%$ & $25\%$ & $50\%$ & $10\%$ & $25\%$ & $50\%$ \\
    \midrule
    \multirow{5}{*}{\rotatebox[origin=c]{90}{\emph{0.8B}}}
    & \textit{Dense}
        & \multicolumn{3}{c}{\dcell{78.0}{72.6}}
        & \multicolumn{3}{c}{\dcell{57.1}{47.6}} \\
    \cmidrule{2-8}
    & Wanda-sp
        & \cell{72.9}{60.2} & \cell{53.6}{30.4} & \cell{7.5}{0.0}
        & \cell{51.9}{50.1} & \cell{35.6}{29.3} & \cell{4.0}{0.0} \\
    & FLAP
        & \cell{74.9}{47.0} & \cell{63.9}{57.0} & \cell{42.6}{34.0}
        & \cell{51.9}{46.8} & \cell{45.8}{43.5} & \cell{33.5}{21.0} \\
    & SlimGPT
        & \cell{74.5}{57.0} & \cell{59.4}{48.1} & \cell{35.9}{13.4}
        & \bcell{55.4}{49.9} & \cell{44.1}{39.6} & \cell{28.3}{9.5}  \\
    & \sk SKIM~(ours)
        & \sk\bcell{76.4}{0.0} & \sk\bcell{71.1}{0.0} & \sk\bcell{53.4}{1.5}
        & \sk\cell{52.1}{0.0} & \sk\bcell{46.9}{0.3} & \sk\bcell{39.3}{0.3} \\
    \midrule
    \multirow{5}{*}{\rotatebox[origin=c]{90}{\emph{2B}}}
    & \textit{Dense}
        & \multicolumn{3}{c}{\dcell{85.9}{79.1}}
        & \multicolumn{3}{c}{\dcell{61.4}{44.6}} \\
    \cmidrule(l){2-8}
    & Wanda-sp
        & \cell{78.9}{54.8} & \cell{63.5}{54.2} & \cell{24.9}{12.9}
        & \bcell{57.6}{56.9} & \cell{47.9}{35.6} & \cell{24.3}{4.8} \\
    & FLAP
        & \bcell{83.8}{82.1} & \cell{75.9}{73.0} & \cell{56.2}{46.5}
        & \cell{56.6}{53.1} & \cell{52.6}{51.1} & \cell{38.1}{35.3} \\
    & SlimGPT
        & \cell{82.0}{74.4} & \cell{73.6}{49.1} & \cell{49.0}{38.1}
        & \cell{56.6}{51.9} & \cell{49.9}{47.1} & \cell{34.1}{26.8} \\
    & \sk SKIM~(ours)
        & \sk\cell{82.9}{7.0} & \sk\bcell{80.8}{0.0} & \sk\bcell{64.5}{0.4}
        & \sk\bcell{57.6}{2.8} & \sk\bcell{54.5}{3.0} & \sk\bcell{48.9}{0.5} \\
    \midrule
    \multirow{5}{*}{\rotatebox[origin=c]{90}{\emph{4B}}}
    & \textit{Dense}
        & \multicolumn{3}{c}{\dcell{88.5}{81.2}}
        & \multicolumn{3}{c}{\dcell{70.2}{67.9}} \\
    \cmidrule{2-8}
    & Wanda-sp
        & \cell{85.9}{82.9} & \cell{70.5}{43.9} & \cell{23.4}{8.2}
        & \cell{58.9}{49.9} & \cell{58.1}{53.6} & \cell{15.0}{0.8} \\
    & FLAP
        & \cell{85.8}{84.1} & \cell{79.9}{76.2} & \cell{57.6}{48.5}
        & \cell{62.4}{58.9} & \cell{60.4}{58.9} & \cell{42.1}{37.3} \\
    & SlimGPT
        & \cell{86.0}{83.1} & \cell{82.4}{80.1} & \cell{61.3}{37.9}
        & \bcell{66.4}{64.7} & \cell{59.6}{49.6} & \cell{41.9}{13.3} \\
    & \sk SKIM~(ours)
        & \sk\bcell{86.1}{20.6} & \sk\bcell{85.0}{44.6} & \sk\bcell{75.1}{0.0}
        & \sk\cell{63.2}{2.0} & \sk\bcell{62.2}{6.8} & \sk\bcell{51.4}{0.0} \\
    \midrule
    \multirow{5}{*}{\rotatebox[origin=c]{90}{\emph{9B}}}
    & \textit{Dense}
        & \multicolumn{3}{c}{\dcell{92.0}{86.1}}
        & \multicolumn{3}{c}{\dcell{75.9}{71.4}} \\
    \cmidrule{2-8}
    & Wanda-sp
        & \cell{87.5}{82.6} & \cell{78.4}{52.5} & \cell{24.1}{5.4}
        & \cell{67.4}{13.8} & \cell{61.2}{20.1} & \cell{21.6}{0.0}  \\
    & FLAP
        & \cell{87.8}{85.5} & \cell{81.9}{75.6} & \cell{48.1}{31.1}
        & \bcell{70.7}{44.0} & \cell{61.9}{23.7} & \cell{39.1}{7.3}  \\
    & SlimGPT
        & \cell{88.6}{87.0} & \bcell{86.4}{82.8} & \cell{59.1}{62.5}
        & \cell{67.7}{33.3} & \cell{58.9}{47.6} & \cell{48.1}{43.1} \\
    & \sk SKIM~(ours)
        & \sk\bcell{90.2}{81.0} & \sk\cell{85.5}{71.0} & \sk\bcell{77.6}{1.3}
        & \sk\cell{65.6}{44.4} & \sk\bcell{66.9}{53.1} & \sk\bcell{58.4}{0.5} \\
    \bottomrule
\end{tabular}}
\end{table}

\subsection{Main results: tool-calling benchmarks}

\paragraph{Quantitative analysis.}
SKIM generally achieves the best structured-decoding accuracy in \cref{tab:main}, with the gap to baselines widening as sparsity increases.
At 50\% sparsity, SKIM retains 68--85\% of dense accuracy, and smaller SKIM-pruned models can match or exceed baseline pruning applied at the next-larger scale.
Free-generation accuracy drops sharply, quite frequently to near zero, as the constrained-distribution distillation objective (\cref{eq:kd-ours}) updates the student's logits only on the valid vocabulary at each step; this is by design for the deployment regime (\S\ref{sec:skim-kd}).
The 9B models at low sparsity are an exception, retaining substantial free-generation accuracy; this is consistent with the larger model having enough remaining capacity to keep free-form behavior alive despite the constrained objective.
We also run SKIM alone on Qwen3.5-27B, without baselines at this scale; \cref{app:seeds} reports its structured accuracy on BFCL-v3 over three seeds.

\begin{figure*}[t]
    \centering
    \includegraphics[width=0.95\textwidth]{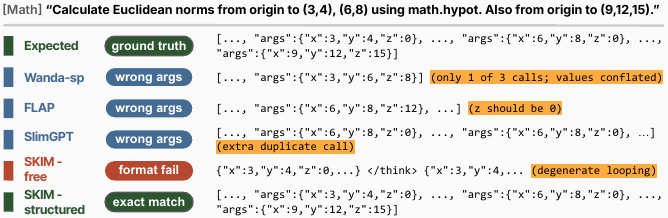}
    \caption{
    BFCL-v3 math example on baseline methods and SKIM. SKIM-free's format errors are intentional: the freed capacity lets SKIM-structured produce the correct call. Baselines fail with wrong arguments or wrong tools, errors grammar cannot fix. See \cref{app:prompts} for more examples.
    }
    \label{fig:qualitative_examples}
\end{figure*}

\paragraph{Qualitative analysis.}
\Cref{fig:qualitative_examples} shows a single BFCL-v3 example (math) comparing baselines, SKIM-free, and SKIM-structured.
The design trade-off behind SKIM is visible: it under-allocates capacity at format positions, so SKIM-free produces malformed JSON, but that same allocation is what lets SKIM-structured pick the correct function and argument once the grammar formatter supplies the surrounding syntax.
Baselines instead distribute capacity uniformly and fail at decision positions, producing well-formed JSON with the wrong tool or argument value.
For the detailed prompt and further examples, we refer readers to \cref{app:prompts}.

\subsection{Cross-model generalization}

\Cref{tab:hammer} repeats the evaluation on
Hammer2.1-3B~\citep{lin2024hammer}, a tool-calling model built on Qwen2.5-Coder. The Qwen3.5 pattern reproduces, with the gap to the strongest baseline widening as sparsity grows: SKIM is competitive at 10\% and 25\% sparsity, and dominates at 50\%.
The cross-backbone agreement supports the claim that SKIM's gains follow from the structure of constrained generation rather than from properties of any specific model architecture.
Hammer also retains substantial free-generation accuracy after SKIM, in contrast to the near-zero collapse on Qwen3.5, because Hammer is already fine-tuned on function-calling data, which leaves its tool-calling priors largely intact.
See \cref{app:hammer_analysis} for a further analysis of the Hammer2.1-3B results.

\begin{table}[t]
\centering
\caption{Post-KD full-match accuracy of Hammer2.1-3B at various sparsity ratios with thinking mode enabled.
Each cell reports \textbf{structured-generation} \textcolor{gray!90}{\scriptsize/\,free-generation} accuracy.
On the tool-specialist model, which is already fine-tuned for function calling, SKIM still achieves the best structured-generation accuracy at 50\% sparsity on both benchmarks.
}
\label{tab:hammer}
\setlength{\aboverulesep}{0pt}
\setlength{\belowrulesep}{0pt}
\setlength{\extrarowheight}{0.15em}
\setlength{\tabcolsep}{0.5em}
\renewcommand{\arraystretch}{1.1}
\resizebox{\textwidth}{!}{\begin{tabular}{>{\centering\arraybackslash}p{0.7cm}p{3.2cm}*{6}{>{\centering\arraybackslash}p{1.6cm}}}
    \toprule
    & & \multicolumn{3}{c}{BFCL-v3} & \multicolumn{3}{c}{API-Bank L1} \\
    \cmidrule(lr){3-5} \cmidrule(lr){6-8}
    & Method~/~Sparsity & $10\%$ & $25\%$ & $50\%$ & $10\%$ & $25\%$ & $50\%$ \\
    \midrule
    \multirow{5}{*}{\rotatebox[origin=c]{90}{\emph{3B}}}
    & \textit{Dense}
        & \multicolumn{3}{c}{\dcell{89.4}{79.9}}
        & \multicolumn{3}{c}{\dcell{76.2}{75.9}} \\
    \cmidrule{2-8}
    & Wanda-sp
        & \cell{88.6}{87.4} & \cell{83.8}{81.0} & \cell{72.2}{65.5}
        & \bcell{76.9}{77.2} & \cell{73.4}{73.9} & \cell{55.4}{54.1} \\
    & FLAP
        & \bcell{89.9}{88.6} & \cell{88.0}{86.1} & \cell{73.0}{68.6}
        & \cell{75.9}{74.9} & \bcell{73.9}{70.7} & \cell{47.3}{30.1} \\
    & SlimGPT
        & \cell{89.4}{87.6} & \cell{85.1}{82.6} & \cell{46.4}{36.1}
        & \cell{75.9}{77.4} & \cell{72.2}{71.4} & \cell{39.1}{30.8} \\
    & \sk SKIM~(ours)
        & \sk\cell{88.8}{83.0} & \sk\bcell{88.1}{80.4} & \sk\bcell{82.8}{69.9}
        & \sk\cell{76.2}{70.9} & \sk\cell{73.4}{66.9} & \sk\bcell{67.7}{14.0} \\
    \bottomrule
\end{tabular}}
\end{table}

\subsection{Agentic reasoning benchmark}

\begin{table}[t]
\begin{minipage}[t]{0.49\linewidth}
\centering
\vspace{0pt}
\caption{Post-KD F1 on HotpotQA at various sparsity ratios with thinking mode enabled. SKIM achieves the best F1 at all sparsity ratios.}
\label{tab:HotpotQA}
\setlength{\aboverulesep}{0pt}
\setlength{\belowrulesep}{0pt}
\setlength{\extrarowheight}{0.15em}
\setlength{\tabcolsep}{0.5em}
\renewcommand{\arraystretch}{1.1}
\scalebox{0.84}{\begin{tabular}{p{3cm}*{3}{>{\centering\arraybackslash}p{1.1cm}}}
\toprule
Method~/~Sparsity & $10\%$ & $25\%$ & $50\%$ \\
\midrule
\textit{Dense}    & \multicolumn{3}{c}{43.6} \\
\midrule
Wanda-sp          & 19.2 & 14.0 & 11.1 \\
FLAP              & 21.6 & 14.5 & 13.3 \\
SlimGPT           & 20.8 & 19.5 & 10.8 \\
\sk SKIM~(ours)   & \sk \textbf{31.7} & \sk \textbf{20.2} & \sk \textbf{14.9} \\
\bottomrule
\end{tabular}}
\end{minipage}\hfill
\begin{minipage}[t]{0.49\linewidth}
\centering
\vspace{0pt}
\captionof{table}{Spider (component F1) and GSM8K (exact match) on Qwen3.5-2B at 10\% sparsity. The baseline prunes with position-uniform saliency and heals with vanilla KD. Each cell reports \textbf{structured} \textcolor{gray!90}{\scriptsize/\,free} performance.}
\label{tab:beyond}
\setlength{\aboverulesep}{0pt}
\setlength{\belowrulesep}{0pt}
\setlength{\extrarowheight}{0.15em}
\setlength{\tabcolsep}{0.5em}
\renewcommand{\arraystretch}{1.1}
\scalebox{0.84}{\begin{tabular}{p{3.6cm}*{2}{>{\centering\arraybackslash}p{1.7cm}}}
\toprule
Method & Spider & GSM8K \\
\midrule
\textit{Dense}        & \dcell{63.5}{16.9} & \dcell{29.5}{14.5} \\
\midrule
Baseline              & \cell{53.1}{16.8}  & \cell{21.0}{22.5} \\
\sk SKIM~(ours)       & \sk\bcell{56.4}{7.8} & \sk\bcell{24.0}{21.5} \\
\bottomrule
\end{tabular}}
\end{minipage}
\end{table}

\Cref{tab:HotpotQA} evaluates Qwen3.5-2B inside a multi-step agent loop on HotpotQA~\citep{yang2018hotpotqa}, where the agent issues retrieval calls and reasons over their outputs.
SKIM achieves the best performance at every sparsity level, consistent with our design: it explicitly retains reasoning capacity in exchange for format capacity, and reasoning is what multi-hop agentic loops most require.
Pruned models still lose substantial performance relative to the dense baseline; this drop reflects a distillation-data mismatch, since we heal on single-turn traces rather than multi-step agent trajectories, and trajectory-level healing closes most of the gap (\cref{app:multistep}).
SKIM nevertheless remains the strongest pruned network at every sparsity ratio, in line with its lead in most settings of \cref{tab:main}.

\subsection{Beyond tool calling}
\label{sec:exp-beyond}

SKIM can be applied to any task decoded under a grammar.
\Cref{tab:beyond} evaluates two further tasks on Qwen3.5-2B at 10\% sparsity: text-to-SQL on Spider~\citep{yu2018spider}, where the SQL keywords and punctuation fixed by the grammar are format positions, and GSM8K~\citep{cobbe2021gsm8k} with each solution written as a JSON object of reasoning steps and a final answer, where the JSON syntax and field names are format positions.
On both tasks SKIM improves the structured metric over the baseline; the free-generation drop is small on GSM8K, where the JSON template is a small part of each output.

\subsection{Ablation studies}
\label{sec:exp-ablation}

\setlength{\intextsep}{2pt}
\setlength{\columnsep}{10pt}
\setlength{\abovecaptionskip}{2pt}
\setlength{\belowcaptionskip}{2pt}
\begin{wrapfigure}{r}{0.47\linewidth}
    \vspace{-0.5em}
    \centering
    \includegraphics[width=\linewidth]{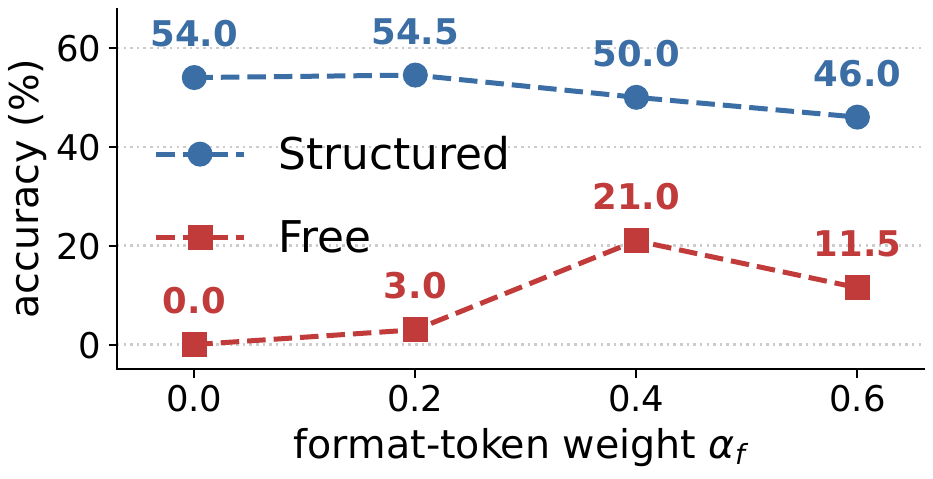}
    \caption{Format-token weight ablation on Qwen3.5-2B (25\% sparsity) with API-Bank L1. Higher $\alpha_f$ trades structured accuracy for free-generation accuracy, as intended.}
    \label{fig:writing-weight-ablation}
\end{wrapfigure}

\paragraph{Category weights.}
\Cref{fig:writing-weight-ablation} sweeps $\alpha_f$, which enters both pruning and hidden-state matching, on Qwen3.5-2B and API-Bank L1 at 25\% sparsity, holding $\alpha_c = \alpha_d = 1.0$ fixed.
Structured accuracy is non-monotonic in $\alpha_f$.
Setting $\alpha_f = 1.0$ weights the three categories equally, so format positions absorb capacity that should go to context and decision; setting $\alpha_f = 0$ discards the signal that format positions still carry for reconstruction.
Structured accuracy peaks near $\alpha_f = 0.2$, matching the finding in \Cref{app:motivating_experiment} that format-specialized capacity is dispensable but not entirely useless.
Free-generation accuracy moves in the opposite direction, rising as $\alpha_f$ increases over most of the sweep, confirming the design intuition that $\alpha_f$ acts as a structured-versus-free trade-off knob.

\setlength{\intextsep}{5pt}
\setlength{\columnsep}{10pt}
\setlength{\abovecaptionskip}{2pt}
\setlength{\belowcaptionskip}{2pt}
\begin{wraptable}{r}{0.47\linewidth}
    \vspace{-0.5em}
    \centering
    \caption{Loss ablation on Qwen3.5-2B (25\% sparsity) with BFCL-v3. Each variant degrades structured accuracy; the
    full SKIM is the best.}
    \label{tab:loss-ablation}
    \setlength{\aboverulesep}{0pt}
    \setlength{\belowrulesep}{0pt}
    \setlength{\extrarowheight}{0.15em}
    \setlength{\tabcolsep}{0.5em}
    \renewcommand{\arraystretch}{1.1}
    \resizebox{\linewidth}{!}{\begin{tabular}{p{3.8cm}*{2}{>{\centering\arraybackslash}p{1.6cm}}}
    \toprule
    Variant & Structured & Free \\
    \midrule
    \sk SKIM (full)
        & \sk \BL{80.8} & \sk 0.0 \\
    \;\;\textcolor{decreasecolor}{($-$)} w/o on-policy.
        & 78.1 & 0.0 \\
    \;\;\textcolor{decreasecolor}{($-$)} w/o feature distill.
        & 79.1 & 0.5 \\
    \;\;\textcolor{increasecolor}{($+$)} w/ vanilla KD.
        & 77.9 & 46.4 \\
    \bottomrule
    \end{tabular}}
\end{wraptable}

\paragraph{Distillation: loss components.}
\Cref{tab:loss-ablation} ablates the components of SKIM's distillation recipe on Qwen3.5-2B and BFCL-v3 data at 25\% sparsity, and \cref{tab:tradeoff} varies the weight of the vanilla output-matching KD when it is added to the recipe.
Removing the on-policy phase causes a substantial drop in structured accuracy because the student is then trained only on ground-truth prefixes, whereas at deployment it must continue from its own, possibly different, decisions.
Removing the feature-distillation term causes a smaller drop, since the hidden-state target adds a coarser but still useful signal at context and decision positions.
Adding the vanilla output-matching KD with coefficient $\beta = 0.5$ causes the largest single drop in structured accuracy but helps retain free-generation accuracy, and the sweep in \cref{tab:tradeoff} makes the structured-vs-free trade-off explicit: pure SKIM has the highest structured accuracy and no free-generation accuracy, vanilla distillation alone the opposite, and intermediate coefficients give operating points between the two.
Driving free-generation accuracy to zero is the right choice only for a specialist served behind a formatter; when one compressed model must also answer unconstrained requests, the coefficient $\beta$, like $\alpha_f$ in pruning, is the knob that sets how much free-generation accuracy is kept.

\paragraph{Context labeling without GPT-4.}
\Cref{tab:labels} replaces the GPT-4 context labels with the rule-based tool-span labeler of \cref{app:pruning-details} and with labeling every prompt token as context; the rule matches GPT-4 labeling on Qwen3.5-2B and trails it by 1.5--4.6 points on Qwen3.5-9B, whereas labeling the full prompt as context degrades accuracy substantially on Qwen3.5-9B.

\begin{table}[t]
\begin{minipage}[t]{0.47\linewidth}
\centering
\vspace{0pt}
\caption{Accuracy on Qwen3.5-2B (25\% sparsity, BFCL-v3) across vanilla-KD coefficients; all rows heal one SKIM-pruned model.}
\label{tab:tradeoff}
\setlength{\aboverulesep}{0pt}
\setlength{\belowrulesep}{0pt}
\setlength{\extrarowheight}{0.15em}
\setlength{\tabcolsep}{0.5em}
\renewcommand{\arraystretch}{1.1}
\scalebox{0.875}{\begin{tabular}{p{3.2cm}*{2}{>{\centering\arraybackslash}p{1.6cm}}}
\toprule
Objective & Structured & Free \\
\midrule
Vanilla KD only                       & 70.0 & 54.0 \\
SKIM, $\beta = 0.50$    & 77.9 & 46.4 \\
SKIM, $\beta = 0.25$    & 78.4 & 34.7 \\
\sk SKIM, $\beta = 0$   & \sk \BL{80.8} & \sk 0.0 \\
\bottomrule
\end{tabular}}
\end{minipage}\hfill
\begin{minipage}[t]{0.47\linewidth}
\centering
\vspace{0pt}
\captionof{table}{Structured accuracy on BFCL-v3 with LLM (GPT-4), tool-span-rule, and full-context labels at 25\% and 50\% sparsity.}
\label{tab:labels}
\setlength{\aboverulesep}{0pt}
\setlength{\belowrulesep}{0pt}
\setlength{\extrarowheight}{0.15em}
\setlength{\tabcolsep}{0.5em}
\renewcommand{\arraystretch}{1.1}
\scalebox{0.875}{\begin{tabular}{lcccc}
\toprule
& \multicolumn{2}{c}{Qwen3.5-2B} & \multicolumn{2}{c}{Qwen3.5-9B} \\
\cmidrule(lr){2-3} \cmidrule(lr){4-5}
Context labels & 25\% & 50\% & 25\% & 50\% \\
\midrule
LLM (default)  & 80.8 & 64.5 & 85.5 & 77.6 \\
Tool-span rule & 81.7 & 64.3 & 84.0 & 73.0 \\
Full context   & 79.8 & 66.8 & 78.0 & 70.0 \\
\bottomrule
\end{tabular}}
\end{minipage}
\end{table}

\subsection{Efficiency}
\label{sec:exp-efficiency}

\setlength{\intextsep}{4pt}
\setlength{\columnsep}{10pt}
\setlength{\abovecaptionskip}{2pt}
\setlength{\belowcaptionskip}{2pt}
\begin{wraptable}{r}{0.42\linewidth}
    \vspace{-0.5em}
    \centering
    \caption{Speedups of SKIM over the dense model at 50\% nominal sparsity (decoding throughput at batch size 256 and prefill speed for 4{,}096 tokens).}
    \label{tab:efficiency}
    \setlength{\aboverulesep}{0pt}
    \setlength{\belowrulesep}{0pt}
    \setlength{\extrarowheight}{0.15em}
    \setlength{\tabcolsep}{0.5em}
    \renewcommand{\arraystretch}{1.1}
    \scalebox{0.875}{\begin{tabular}{p{2.4cm}*{2}{>{\centering\arraybackslash}p{1.5cm}}}
    \toprule
    Model & Decoding & Prefill \\
    \midrule
    Qwen3.5-2B  & 1.29$\times$ & 1.23$\times$ \\
    Qwen3.5-9B  & 1.41$\times$ & 1.37$\times$ \\
    Qwen3.5-27B & 1.59$\times$ & 1.51$\times$ \\
    \bottomrule
    \end{tabular}}
\end{wraptable}

SKIM physically removes MLP channels and attention components, so the pruned checkpoint is a smaller dense model that runs without sparse kernels.
\Cref{tab:efficiency} reports decoding throughput and prefill speed relative to the dense model at 50\% nominal sparsity, measured on one 48\,GB A6000 for the 2B and 9B models and on two for the 27B model.
Width pruning keeps the sequential depth, so single-request speedups are modest; the gains grow with model size and batch size, reaching 1.59$\times$ throughput at batch size 256 for the 27B model, and \cref{app:throughput} reports batch size one as well.

%% file: main/discussion.tex
\section{Conclusion}
\label{sec:discussion}

We have presented SKIM, a pruning-and-distillation framework for
LLM agents deployed under structured generation. SKIM operates from
the observation that tokens in a tool-calling output do not all play
the same role at deployment: a pruned model that preserves
predictive fidelity uniformly across positions wastes capacity on
outputs the constraint mechanism already guarantees. SKIM exploits a
context/decision/format partition of token positions in both stages,
with a category-aware pruning saliency derived from a heteroscedastic
surrogate precision and a distillation objective that minimizes
forward KL on the constrained student distribution with
position-aware hidden-state matching. Across the Qwen3.5 series and a Hammer-2.1-3B backbone, SKIM generally achieves the best structured-decoding accuracy on BFCL-v3 and API-Bank, with the largest gains at 50\% sparsity where existing methods struggle the most.

\paragraph{Future work and limitations.}
We discuss the limitations of our work and outline directions for future work.
We evaluated SKIM on models up to 27B; validation at larger scales and on non-Qwen families remains future work, and the larger-compute regime may surface trends invisible at our current scale.
Our distillation corpus consists of single-turn function-calling traces, so SKIM recovers multi-step agentic behavior only indirectly; distillation on multi-step agentic trajectories largely closes the HotpotQA gap.
The pruning stage adopts a layer-wise reconstruction framework, leaving category-aware extensions to end-to-end pruning methods~\citep{ma2023llm,xia2023sheared}, and to calibration-based quantization~\citep{frantar2022gptq,lin2024awq} open.
More broadly, SKIM instantiates a general principle that constrained inference changes which output positions the model controls and therefore which weights compression should preserve, with applications to code generation and other constrained-output settings.

%% file: appendix/related_works.tex
\section{Further Related Works}
\label{app:further_related_works}

\paragraph{Tool-calling LLM agents.}
Large language models are widely deployed as autonomous agents interacting with external APIs \citep{yao2022react, schick2023toolformer, patil2024gorilla, qin2023toolllm, zhang2025xlam, li2023api}. As these agents transition to production-scale deployments, they face intertwined bottlenecks in efficiency and reliability. While prior work improved robustness via tool embeddings \citep{hao2023toolkengpt} or function masking \citep{lin2024hammer}, reliable deployment fundamentally requires syntactically flawless API requests (e.g., strict JSON). Because even minor structural deviations cause cascading execution failures, production pipelines overwhelmingly rely on mechanisms that explicitly enforce the desired output structure.

\paragraph{Structured generation.}
Structured generation enforces a grammar at inference time, guaranteeing structurally valid outputs.
Early constrained-decoding algorithms~\citep{deutsch2019general,geng2023grammar} formalized this idea, and modern engines~\citep{beurer2023prompting, willard2023efficient, lew2023sequential} realize it efficiently by masking invalid tokens against predefined schemas or finite-state machines.
Their objective is faithful, low-overhead grammar enforcement, and the LLM is treated as a black box that the engine wraps without modification.
A grammar, however, makes some token positions deterministic and others not - at deterministic positions the constraint uniquely fixes the next token, while at the rest the model's prediction selects the output.
This position-level asymmetry is outside the scope of grammar-enforcement engines, but for compression methods designed for the structured-generation regime it is exactly the signal worth exploiting for compression under structured generation.

\paragraph{LLM pruning.}
LLM pruning methods divide along two axes: pruning granularity and saliency framework.
Along the granularity axis, methods range from unstructured pruning, which scores individual weights~\citep{frantar2023sparsegpt,sun2023simple}, to structured pruning that removes whole architectural units (channels, attention
heads, or transformer blocks) for wall-clock speedups on dense hardware~\citep{ma2023llm,ashkboos2024slicegpt,men2025shortgpt,an2024fluctuation,ling2024slimgpt}.
Along the saliency-framework axis, magnitude-based scoring~\citep{han2015deep_compression} and gradient-based scoring~\citep{ma2023llm,molchanov2019taylor} estimate per-unit importance from weights and gradients.
Layer-wise reconstruction, the framework most relevant to SKIM, scores units by how much zeroing them perturbs the layer's output on calibration activations~\citep{frantar2023sparsegpt,sun2023simple, an2024fluctuation,ling2024slimgpt}; within this family, SparseGPT and SlimGPT additionally update the remaining weights to absorb the reconstruction error, while Wanda and FLAP do not.
After pruning, most methods recover lost capability through post-hoc fine-tuning or knowledge distillation~\citep{xia2023sheared,muralidharan2024compact}.
SKIM extends layer-wise reconstruction by partitioning calibration activations by the role each position plays under structured generation, and pairs the position-aware saliency with a constrained-distribution distillation recipe rather than standard output-matching distillation.
We accordingly select Wanda~\citep{sun2023simple}, FLAP~\citep{an2024fluctuation}, and SlimGPT~\citep{ling2024slimgpt} as baselines, spanning layer-wise reconstruction and providing the most direct comparison to SKIM.

\paragraph{LLM distillation.}
Knowledge distillation (KD) facilitates knowledge transfer from a teacher to a student model by matching output distributions or intermediate hidden states~\citep{hinton2015distilling,romero2014fitnets}.
While KD has long been a well-established paradigm in language modeling~\citep{jiao2020tinybert,sanh2019distilbert,sun2019patient}, recent research has shifted toward on-policy approaches, where students learn from their own generated trajectories to improve alignment in generative tasks~\citep{gu2024minillm,agarwal2024policy,yang2025qwen3}.
Additionally, KD has been utilized to recover performance degradation following LLM pruning~\citep{muralidharan2024compact,sreenivas2024llm}.
However, to the best of our knowledge, no existing studies have explored KD specifically within the context of grammar-constrained decoding.

%% file: appendix/method_details.tex
\section{Implementation Details}
\label{app:implementation_details}

\subsection{SKIM: Pruning}
\label{app:pruning-details}

\paragraph{Per-category normalization.}
In practice, the per-category activation sums in the saliency
formulation are averaged within each category before
$\alpha$-weighted aggregation. The implemented saliency score is
\begin{equation}\label{eq:context_aware_saliency_score_normalized}
    S_i^{\text{pos}}
    \;=\; \Bigg(
        \alpha_c \cdot \frac{1}{|P_c|} \sum_{t \in P_c} X_{ti}^2
        + \alpha_d \cdot \frac{1}{|P_d|} \sum_{t \in P_d} X_{ti}^2
        + \alpha_f \cdot \frac{1}{|P_f|} \sum_{t \in P_f} X_{ti}^2
    \Bigg) \cdot \big\| W_{i,:} \big\|_2^2,
\end{equation}
which differs from \cref{eq:context_aware_saliency_score} by the
$1/|P_\kappa|$ factors. Without this normalization, the most populous
category (typically format) would dominate the saliency irrespective
of the chosen $\alpha$ values: even with $\alpha_f \ll \alpha_d$, an
unweighted format sum can outweigh a small but heavily weighted
decision sum simply because $|P_f| \gg |P_d|$ on the calibration set.
The same per-category averaging is used in the self-attention and
linear-attention saliencies described below.

\paragraph{Surrogate precision design: additive vs.\ exponential.}
The form of $\tilde{H}_{ii}^{\text{pos}}$ is one of several ways to
encode the position-level inductive bias into the diagonal Hessian.
The form used throughout SKIM is the \emph{additive} one already
introduced in
\cref{eq:context_aware_saliency_score_normalized}: each category
contributes a non-negative weighted average of squared activations
into the precision. A natural alternative is an \emph{exponential}
form that treats context and decision positions as additive precision
gains and format positions as multiplicative variance injection:
\begin{equation}\label{eq:exp-saliency}
S_i^{\text{exp}}
\;=\; \Bigg(
    \alpha_c \cdot \tfrac{1}{|P_c|} \textstyle\sum_{t \in P_c} X_{ti}^2
    + \alpha_d \cdot \tfrac{1}{|P_d|} \textstyle\sum_{t \in P_d} X_{ti}^2
\Bigg)
\cdot \exp\!\Bigg(\!\!-\,2 \alpha_f \cdot \tfrac{1}{|P_f|}
\textstyle\sum_{t \in P_f} X_{ti}^2 \Bigg)
\cdot \big\| W_{i,:} \big\|_2^2.
\end{equation}
This decouples format positions from the additive precision pool and
instead penalizes neurons whose activations concentrate on format
tokens through an exponential gate.

\Cref{tab:saliency_design} compares the two on Hammer-3B at 20\%
sparsity. The additive form is both simpler and stronger on both
benchmarks, so we adopt it for all main results.

\begin{table}[t]
\centering
\small
\caption{
Surrogate precision design comparison on Hammer-3B at 20\% sparsity. Numbers are tool-use accuracy (\%) after pruning and distillation under matched compute.
}
\label{tab:saliency_design}
\begin{tabular}{lcc}
\toprule
Surrogate precision & BFCL-v3 & API-Bank \\
\midrule
Exponential (\cref{eq:exp-saliency}) & 80.3 & 63.7 \\
Additive (\cref{eq:context_aware_saliency_score_normalized})
  & \textbf{88.1} & \textbf{72.9} \\
\bottomrule
\end{tabular}
\end{table}

\paragraph{Pruning unit and saliency by module type.}
SKIM applies category-aware saliency at three different
granularities, chosen to match the architecture of each module
type in Qwen3.5. We describe the prunable unit
(intermediate neuron, KV group, or linear-attention
group); the saliency in \cref{eq:context_aware_saliency_score_normalized}
is applied to the input feature of the projection whose input
feature index coincides with that unit.

\begin{itemize}
    \item \textbf{MLP}. The structured unit is the intermediate MLP
    neuron, which appears as an output dimension of
    \texttt{gate\_proj} and \texttt{up\_proj} and as an input
    dimension of \texttt{down\_proj}. We score neuron $j$ by
    applying \cref{eq:context_aware_saliency_score_normalized}
    to \texttt{down\_proj}, identifying its input feature index
    with the neuron index. Removing neuron $j$ zeros the
    corresponding output dimension of \texttt{gate\_proj} and
    \texttt{up\_proj} and the corresponding input dimension of
    \texttt{down\_proj}.

    \item \textbf{Self-attention} (regular softmax attention,
    with \texttt{q\_proj}, \texttt{k\_proj}, \texttt{v\_proj},
    \texttt{o\_proj}). The structured unit is the KV group: the
    head dimensions associated with one shared $(K, V)$ pair
    across the query heads in that group.
    \cref{eq:context_aware_saliency_score_normalized} is applied
    per KV group rather than per \texttt{o\_proj} channel, because
    the group, not the channel, is the pruning unit and its query
    heads share one $(K, V)$ pair: we score each KV group
    by the projected head-output energy
    $\| A_t^{(h)} W_O^{(h)} \|_2^2$, summed over the query heads
    in the group and averaged within each token category, then
    aggregated across categories with the same
    $(\alpha_c, \alpha_d, \alpha_f)$ as MLP pruning. Dropping a
    KV group zeros the corresponding output dimensions of
    $W_Q, W_K, W_V$ and the corresponding input dimensions of
    $W_O$ in that group.

    \item \textbf{Linear attention} (the \texttt{linear\_attn}
    module in Qwen3.5's Gated DeltaNet block, architecturally
    distinct from regular self-attention). The structured unit
    is the linear-attention group: the channel block shared
    across the input projections, \texttt{conv1d}, and
    \texttt{out\_proj}. We score individual channels by applying
    \cref{eq:context_aware_saliency_score_normalized} to
    \texttt{out\_proj} (identifying its input feature index with
    the channel index), then aggregate channel scores within
    each group; the group, not the channel, is the prunable
    unit. Dropping a group zeros the corresponding output
    dimensions of the input projections (and \texttt{conv1d}, if
    present) and the corresponding input dimensions of
    \texttt{out\_proj} in the same layer.
\end{itemize}

\paragraph{Token category labels on calibration data.}
We classify calibration positions into context, decision, and format
using a hybrid rule.
\emph{Output positions} (positions the model generates) are labeled
by their role in the call: function names, argument names and values,
and the brackets and commas that set the number of calls are decision
positions unless the grammar forces them ($|\mathcal{V}_t| = 1$); all
other output positions are format. $|\mathcal{V}_t|$ alone does not
separate the two, as the JSON grammar admits optional whitespace almost
everywhere.
\emph{Input positions} are not all treated as context: only
\emph{tool-relevant} input tokens (function signatures, argument
descriptions, the user query) are labeled context, and these labels
are produced by prompting GPT-4~\citep{openai2024gpt4technicalreport} over each calibration prompt and
asking which spans are tool-relevant. Remaining input tokens (system
boilerplate, role markers, special tokens) are excluded from
$P_c, P_d, P_f$ entirely and do not contribute to any saliency sum.

\paragraph{Context labels without GPT-4.}
\Cref{tab:labels} compares the GPT-4 labels with two alternatives that need no proprietary model: a tool-span rule that labels as context the spans holding tool or function names, parameter names, parameter descriptions, and the user query, which selects about 26\% of the prompt tokens, and labeling every prompt token as context.
The rule matches GPT-4 labeling on Qwen3.5-2B and trails it by 1.5--4.6 points on Qwen3.5-9B, whereas labeling the full prompt as context degrades accuracy substantially on Qwen3.5-9B, so identifying the relevant spans matters more than preserving every token of the prompt, and the rule is a reproducible alternative.

\paragraph{Hyperparameters and procedure.}
For all main results we use $\alpha_c = 1.0$, $\alpha_d = 1.0$,
$\alpha_f = 0.2$, held fixed across model and target sparsity. The
calibration set is the same xLAM-FC~\citep{liu2024apigen} subset used
by the layer-wise reconstruction baselines. Pruning is one-shot in
the Wanda sense~\citep{sun2023simple}: for each module we compute the
saliency once, prune the lowest-scoring units to that module's target
sparsity, and do not perform sequential weight updates as in
SparseGPT~\citep{frantar2023sparsegpt}. The per-layer sparsity
allocation across the network follows
FLAP~\citep{an2024fluctuation}; SKIM's contribution is in the
saliency, not in this allocation: \cref{tab:allocation} shows that
giving every layer the same budget performs comparably.

\begin{center}
\small
\captionof{table}{Pruning schedule on Qwen3.5-2B and BFCL-v3 (three further seeds, mean $\pm$ std.).}
\label{tab:allocation}
\begin{tabular}{lcc}
\toprule
Schedule & 25\% sparsity & 50\% sparsity \\
\midrule
Adaptive across layers (default) & 79.8 $\pm$ 1.7 & 64.8 $\pm$ 0.9 \\
Uniform across layers            & 78.7 $\pm$ 0.9 & 66.7 $\pm$ 0.7 \\
\bottomrule
\end{tabular}
\end{center}

\paragraph{Sensitivity to the category weights.}
\Cref{tab:weight-sensitivity} compares three weightings of $(\alpha_c, \alpha_d, \alpha_f)$ on a random subset of 100 BFCL-v3 items, so the values are comparable only within the table.
The context- and decision-heavy weighting is best at every scale and sparsity, followed by the uniform weighting and then the format-heavy weighting (the two tie at 25\% on Qwen3.5-2B); the key choice is to prioritize context and decision positions over format positions rather than one specific tuple.

\begin{center}
\small
\captionof{table}{Structured accuracy on a 100-item subset of BFCL-v3 under three category weightings.}
\label{tab:weight-sensitivity}
\begin{tabular}{lcccc}
\toprule
& \multicolumn{2}{c}{Qwen3.5-2B} & Qwen3.5-4B & Qwen3.5-9B \\
\cmidrule(lr){2-3} \cmidrule(lr){4-4} \cmidrule(lr){5-5}
Weighting $(\alpha_c, \alpha_d, \alpha_f)$ & 25\% & 50\% & 50\% & 50\% \\
\midrule
Context- and decision-heavy $(2.0, 2.0, 0.5)$ & \textbf{80.0} & \textbf{69.0} & \textbf{79.0} & \textbf{81.0} \\
Uniform $(1.0, 1.0, 1.0)$                     & 79.0 & 54.0 & 75.0 & 77.0 \\
Format-heavy $(0.5, 0.5, 2.0)$                & 79.0 & 47.0 & 73.0 & 74.0 \\
\bottomrule
\end{tabular}
\end{center}

\subsection{SKIM: Distillation}
\label{app:kd-details}

\paragraph{Loss.}
The full SKIM training objective combines the on-policy
constrained-KL term (\cref{eq:kd-ours}) with a position-aware
hidden-state matching term:
\begin{equation}\label{eq:skim-loss-app}
    \mathcal{L}_{\text{SKIM}}
    \;=\; \mathcal{L}_{\text{KD}}
    \;+\; \hat{\lambda}_{H} \, \mathcal{L}_{\text{hidden}},
\end{equation}
where
\begin{equation}\label{eq:hidden-loss}
    \mathcal{L}_{\text{hidden}}
    \;=\; \sum_{\ell \in L} \sum_{t}
    \hat{\alpha}_{\mathrm{cat}(t)} \,
    \big\| h_t^{(p, \ell)} - h_t^{(q, \ell)} \big\|_2^2,
    \qquad
    \hat{\alpha}_{c} \;=\; \frac{\alpha_{c}}{\sum_{c'}\alpha_{c'}},
\end{equation}
and $h_t^{(p, \ell)}, h_t^{(q, \ell)}$ are the teacher and student
hidden states at position $t$ following transformer block $\ell$. The
matched layer set $L$ contains three transformer block outputs spaced
every other layer, i.e., $L = \{\ell_0, \ell_0 + 2, \ell_0 + 4\}$ for
a chosen $\ell_0$. The per-position weights $\hat{\alpha}_{\mathrm{cat}(t)}$
form a probability distribution over the three categories
($\sum_{c} \hat{\alpha}_c = 1$), so $\hat{\alpha}$ controls only the
\emph{relative} importance of context, decision, and format positions
while the scalar $\hat{\lambda}_H$ controls the overall weight of the
hidden-state objective. The unnormalized weights
$(\alpha_c, \alpha_d, \alpha_f) = (1.0, 1.0, 0.2)$ are reused exactly
from the pruning saliency, giving
$(\hat{\alpha}_c, \hat{\alpha}_d, \hat{\alpha}_f) = (5/11, 5/11, 1/11)$;
we use $\hat{\lambda}_H = 1.1$ throughout, and we match raw L2
distances without per-layer normalization or cosine terms.

\paragraph{On-policy distillation (OPD).}
Off-policy distillation evaluates both models only on ground-truth prefixes $y^\ast_{<t}$.
Under structured decoding, a decision that differs from the ground truth, such as another function name, makes the grammar follow that function's schema, so the following positions and valid sets $\mathcal{V}_t$ do not appear in the ground-truth sequence.
A pruned student often makes such decisions and must continue from them at deployment.
On-policy distillation~\citep{agarwal2024policy,gu2024minillm} trains the student on its own responses; SKIM generates them under the grammar, so the constrained KL of \cref{eq:kd-ours} covers the decision positions and valid sets the deployed student reaches.

\textit{Phase 1 -- off-policy warm-up (first 25\% of optimizer steps).}
Targets use the ground-truth completions $y^\ast$ from the
xLAM-FC subset: the teacher distribution
$p_\phi(\cdot \mid x, y^\ast_{<t})$ is evaluated by teacher-forcing the
ground-truth $y^\ast_{<t}$, and the student matches this
distribution under $\mathcal{L}_{\text{KD}}$ and
$\mathcal{L}_{\text{hidden}}$ at the same positions; this warm-up gives stable targets before the student sees its noisy early rollouts.

\textit{Phase 2 -- on-policy (remaining 75\% of optimizer steps).}
For each $x$, the current student generates $y$ of $\mathcal{L}_{\text{KD}}$ (up to 300 new tokens) greedily from $q_\phi$ instead of sampling, as at evaluation, under the example's JSON-schema grammar (XGrammar), with end-of-sequence masked until the grammar can terminate; an example whose grammar fails to compile or whose rollout is empty falls back to its ground-truth completion.
Categories and $\mathcal{V}_t$ are recomputed on the rollout, where the teacher is teacher-forced and both losses are computed; no gradient flows through it (the $\texttt{sg}$ in $\mathcal{L}_{\text{KD}}$).

\paragraph{Hyperparameters.}
All models share the same training hyperparameters. Training runs for 2 epochs at a learning rate of $1 \times 10^{-5}$, with effective batch size 4 (4 gradient accumulation steps) and maximum sequence length 1{,}300 tokens. Distillation uses a 4096-sample subset of xLAM-FC~\citep{liu2024apigen}, matching the subset used for pruning calibration. The distillation temperature is 2.0, applied to both teacher and student logits before computing $D_{\text{KL}}(p_\phi \,\|\, q_\phi)$. We run all experiments on NVIDIA A6000 (48~GB VRAM), A100 (80~GB VRAM), and H200 (141~GB VRAM) GPUs, with 256~GB system RAM and 1~TB local NVMe storage per node. With two A100 GPUs, we can run a single SKIM experiment which takes around 6--12 hours end to end, depending on the model size.

\paragraph{Mixing with vanilla distillation.}
\Cref{tab:tradeoff} adds the vanilla output-matching distillation of \cref{eq:kd-vanilla} to the SKIM objective with a coefficient $\beta$ on Qwen3.5-2B at 25\% sparsity.
Pure SKIM has the highest structured accuracy and no free-generation accuracy, vanilla distillation alone the opposite, and intermediate coefficients give operating points in between.

%% file: appendix/further_experiments.tex
\section{Further experiments}
\label{app:further_experiments}

\subsection{Motivating experiment}
\label{app:motivating_experiment}

To complement the post-hoc error analysis in \Cref{fig:intro}~(right),
we run a controlled ablation on MLP neurons. We restrict the
experiment to MLP pruning, since MLP intermediate channels admit a
clean per-neuron specialization score; the resulting evidence is
therefore specifically about the MLP pathway. For each MLP
intermediate neuron $i$ in each transformer layer, we measure its
mean squared activation on tokens of each category
$c \in \{\text{context}, \text{decision}, \text{format}\}$ over the
calibration set:
\begin{equation}
    A_i^c \;=\; \frac{1}{|\mathcal{T}_c|}
    \sum_{t \in \mathcal{T}_c} h_i(t)^2,
\end{equation}
where $h_i(t)$ is the $i$-th intermediate activation at token
position $t$ and $\mathcal{T}_c$ is the set of calibration positions
of category $c$. The category-$c$ specialization score of neuron $i$
is then
\begin{equation}
    s_i^c \;=\; \frac{A_i^c}{\sum_{c'} A_i^{c'}},
\end{equation}
and we call neuron $i$ \emph{category-$c$ specialized} if
$s_i^c \geq \theta$ with $\theta = 0.6$, i.e., more than 60\% of its
activation budget on the calibration set concentrates on a single
category. For each category $c$, we then zero the corresponding
down-projection rows for all category-$c$-specialized neurons
across all layers. \emph{random} ablates an equal number of
random MLP neurons; \emph{dense} is the unpruned model.

\begin{figure}[t]
    \centering
    \includegraphics[width=0.95\linewidth]{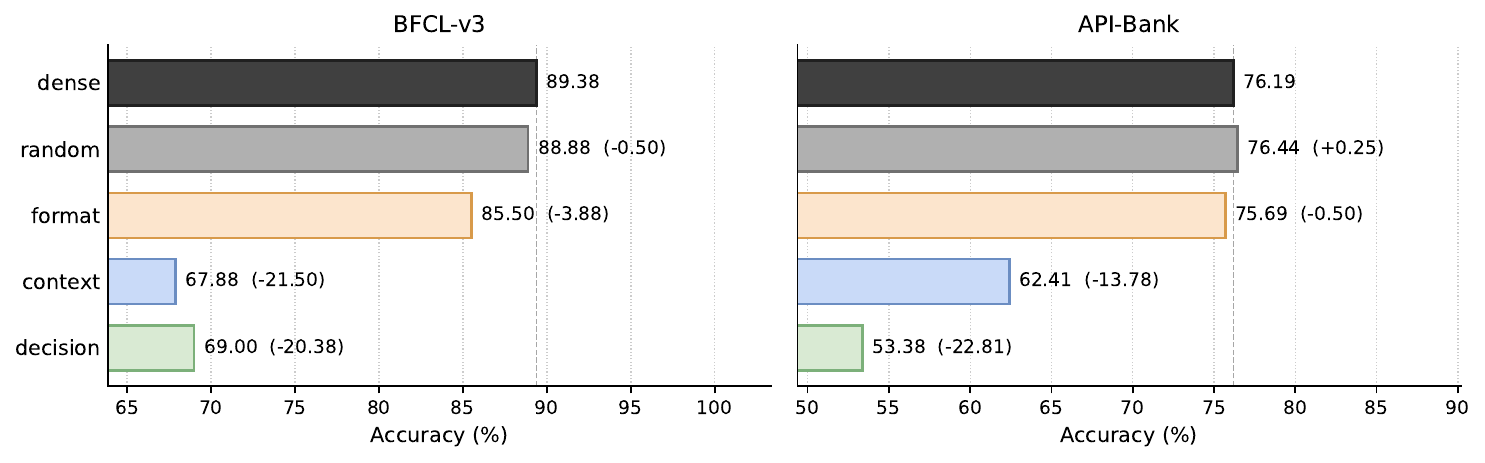}
    \caption{Task accuracy under category-specialized MLP ablation
    on BFCL-v3 (left) and API-Bank L1 (right) on Hammer2.1-3B model. \emph{context},
    \emph{decision}, and \emph{format} ablate MLP neurons whose
    activation budget on the calibration set concentrates
    ($s_i^c \geq 0.6$) on the corresponding token category;
    \emph{random} ablates the same number of MLP neurons uniformly
    at random; \emph{dense} is the unpruned baseline. Format-specialized
    ablation barely affects accuracy, while context- or
    decision-specialized ablation causes sharp drops in accuracy relative to the dense baseline.}
    \label{app:fig:motivation_ablation}
\end{figure}

\Cref{app:fig:motivation_ablation} reports task accuracy on BFCL-v3 and
API-Bank. Ablating format-specialized neurons keeps accuracy within
$\sim$4 points of random on both benchmarks, while ablating
context-specialized neurons drops accuracy by 13.8 to 21.5 points
and ablating decision-specialized neurons drops accuracy by 20.4 to
22.8 points. These results provide direct evidence that
format-specialized MLP capacity is largely dispensable when
structured decoding handles formatting, whereas
context- and decision-specialized MLP capacity is essential for task
accuracy and must be preserved by pruning.

\begin{figure}[t]
    \centering
    \includegraphics[width=0.5\linewidth]{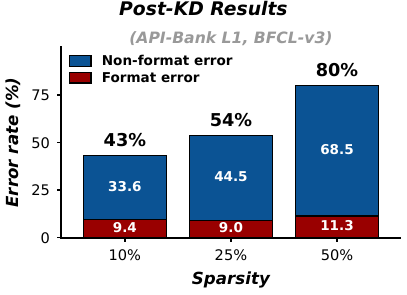}
    \caption{
    Pruned-then-distilled model performance averaged over three baselines, two benchmarks, and Qwen3.5-0.8B/2B (\cref{sec:motivation}). Pruned baseline error grows with sparsity, but non-format error dominates: capacity is misallocated to format behavior the constraint mechanism enforces for free.
    }
    \label{app:fig:error_decomposition}
\end{figure}

\subsection{Pruning analysis for tool-expert model}
\label{app:hammer_analysis}

If SKIM's position-aware advantage exploits role specialization at
the neuron level, one expects more such specialization in a model
that has been fine-tuned specifically for tool calling than in a
general-purpose model. Hammer-2.1-3B is a tool-specifically
fine-tuned LLM in the same parameter range as Qwen3.5-2B, and we
contrast the two directly using the specialization criterion of
\S\ref{app:motivating_experiment}, namely $s_i^c \geq \theta = 0.6$ for at least one
token category $c$.

\begin{figure}[t]
\centering
\includegraphics[width=0.45\linewidth]{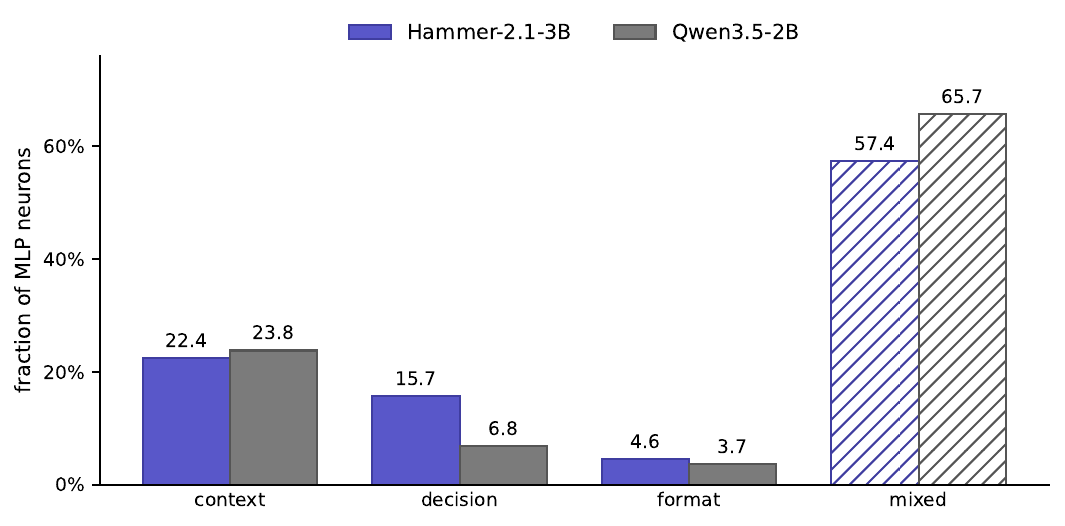}
\caption{Aggregate fraction of MLP neurons specialized for each
token category on Hammer2.1-3B and Qwen3.5-2B. \emph{context},
\emph{decision}, \emph{format} correspond to context, decision, and
format positions respectively; \emph{mixed} denotes neurons that do
not satisfy the $\theta = 0.6$ threshold for any
single category.}
\label{fig:specialization-aggregate}
\end{figure}

\Cref{fig:specialization-aggregate} reports the aggregate fraction
of MLP neurons specialized for each category across the entire
network. The most striking gap is in the decision category: 15.7\%
of Hammer-2.1-3B's MLP neurons are decision-specialized, against
only 6.8\% in Qwen3.5-2B. Most of the difference comes out of the
mixed pool (57.4\% in Hammer vs.\ 65.7\% in Qwen3.5-2B), with
context- and format-specialized fractions roughly comparable across
the two models. The tool-specific fine-tuning regime appears to
convert mixed neurons into decision-specialized ones, consistent
with the intuition that a function-calling objective rewards neurons
that respond selectively at constrained branch points, where the format admits more than one continuation.

\begin{figure}[t]
\centering
\includegraphics[width=0.5\linewidth]{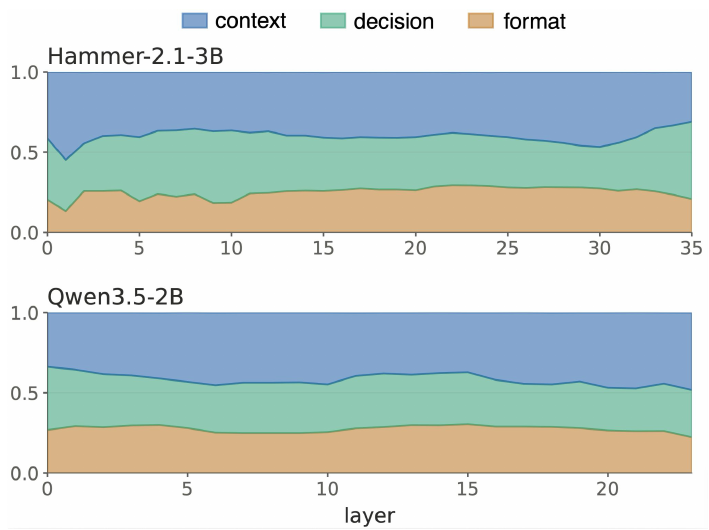}
\caption{Layer-wise fraction of MLP neurons assigned to each
token category by their largest selectivity score on Hammer-2.1-3B
(top) and Qwen3.5-2B (bottom); unlike \cref{fig:specialization-aggregate},
every neuron is assigned to one category. At most layers, Hammer-2.1-3B
has the larger decision fraction.}
\label{fig:specialization-layerwise}
\end{figure}

\Cref{fig:specialization-layerwise} resolves this picture by depth.
At most layers beyond the first few, Hammer2.1-3B sits above
Qwen3.5-2B on the decision share, recapitulating the aggregate
pattern at the layer level.

The implication for SKIM is that the position-aware saliency has a
larger pool of clearly specialized neurons to work with on
tool-expert models. This is consistent with
\Cref{tab:hammer}: SKIM's margin over baselines on Hammer2.1-3B is
largest at 50\% sparsity, where the budget is tight enough that
correctly identifying which neurons can be removed (the
format-specialized ones) and which must be preserved (the
context- and decision-specialized ones) determines the outcome.

\subsection{Memory Efficiency}
\label{app:efficiency}

We report SKIM's memory footprint on Qwen3.5-2B, measured on physically compacted models in which the masked neurons and groups have been removed from the weight tensors.
\Cref{tab:eff} reports parameter count, whole-model sparsity, and GPU memory at two points: \emph{at load} (after the compacted model has been loaded onto the GPU) and \emph{peak} (the maximum observed during a representative inference run).
\emph{Whole-model sparsity} is the realized parameter reduction across the full model:
the nominal setting applies only to the prunable projections (MLP and self/linear-attention modules), so embeddings, lm-head, and untouched layers pull the realized number down.
At 50\% nominal sparsity, SKIM removes 34.14\% of the parameters, which reduces both the load-time and the peak GPU memory by roughly 1.2\,GiB at both points.

\begin{table}[t]
\centering
\small
\caption{Memory footprint of Qwen3.5-2B under SKIM, measured on
physically compacted models. \emph{At load} is the GPU memory after
weight loading; \emph{peak} is the maximum GPU memory observed at any
point during one representative inference run.}
\label{tab:eff}
\setlength{\tabcolsep}{8pt}
\begin{tabular}{lrrrr}
\toprule
          & Params  & Whole-model & GPU memory & Peak GPU   \\
          &         & sparsity    & at load    & memory     \\
\midrule
Dense     & 1.88\,B & --          & 7.010\,GiB & 7.358\,GiB \\
SKIM-10\% & 1.75\,B &  6.77\%     & 6.792\,GiB & 7.130\,GiB \\
SKIM-25\% & 1.56\,B & 17.07\%     & 6.420\,GiB & 6.754\,GiB \\
SKIM-50\% & 1.24\,B & 34.14\%     & 5.821\,GiB & 6.151\,GiB \\
\bottomrule
\end{tabular}
\end{table}

\subsection{Throughput}
\label{app:throughput}

\Cref{tab:throughput} reports decoding throughput and prefill speed relative to the dense model at 50\% nominal sparsity, measured on one 48\,GB A6000 for the 2B and 9B models and on two for the 27B model.
Width pruning keeps the sequential depth, so single-request speedups are modest, and the gains grow with model size and with batch size, as expected for width pruning.

\begin{table}[t]
\centering
\small
\caption{Speedups over the dense model at 50\% nominal sparsity.}
\label{tab:throughput}
\begin{tabular}{lcccc}
\toprule
Model & Parameters & Decode, batch 1 & Decode, batch 256 & Prefill, 4{,}096 tokens \\
\midrule
Qwen3.5-2B  & 1.88B $\to$ 1.24B & 1.03$\times$ & 1.29$\times$ & 1.23$\times$ \\
Qwen3.5-9B  & 8.95B $\to$ 6.06B & 1.13$\times$ & 1.41$\times$ & 1.37$\times$ \\
Qwen3.5-27B & 26.9B $\to$ 16.7B & 1.17$\times$ & 1.59$\times$ & 1.51$\times$ \\
\bottomrule
\end{tabular}
\end{table}

\subsection{Seed variation and larger scale}
\label{app:seeds}

\Cref{tab:seeds-methods} repeats the Qwen3.5-2B comparison of \cref{tab:main} on BFCL-v3 over three further random seeds, and \cref{tab:seeds-skim} repeats SKIM on Qwen3.5-4B and 9B and adds Qwen3.5-27B, for which no baselines were run.
In \cref{tab:main} the free-generation accuracy of SlimGPT on Qwen3.5-9B at 50\% sparsity (62.5) slightly exceeds its structured accuracy (59.1); over three seeds the two values are tied within variation (structured 60.0 $\pm$ 3.1, free 60.1 $\pm$ 4.8).
Constrained decoding changes the token space available to the model rather than correcting its prediction, so it can occasionally exclude a semantically acceptable output of unconstrained decoding.

\begin{center}
\small
\captionof{table}{Structured accuracy of each method on Qwen3.5-2B and BFCL-v3 over three further random seeds. Each cell reports the mean $\pm$ standard deviation across the seeds.}
\label{tab:seeds-methods}
\begin{tabular}{lcc}
\toprule
Method & 25\% sparsity & 50\% sparsity \\
\midrule
Wanda-sp & 64.9 $\pm$ 1.0 & 26.4 $\pm$ 1.3 \\
SlimGPT  & 74.9 $\pm$ 0.7 & 47.8 $\pm$ 0.4 \\
FLAP     & 76.8 $\pm$ 0.7 & 54.7 $\pm$ 0.6 \\
SKIM (ours) & \textbf{79.8 $\pm$ 1.7} & \textbf{64.8 $\pm$ 0.9} \\
\bottomrule
\end{tabular}
\end{center}

\begin{center}
\small
\captionof{table}{Structured accuracy of SKIM on BFCL-v3 over three further random seeds (mean $\pm$ std.).}
\label{tab:seeds-skim}
\begin{tabular}{lcc}
\toprule
Model & 25\% sparsity & 50\% sparsity \\
\midrule
Qwen3.5-2B  & 79.8 $\pm$ 1.7 & 64.8 $\pm$ 0.9 \\
Qwen3.5-4B  & 84.7 $\pm$ 1.2 & 76.3 $\pm$ 0.6 \\
Qwen3.5-9B  & 84.7 $\pm$ 0.6 & 77.3 $\pm$ 3.5 \\
Qwen3.5-27B & 89.0 $\pm$ 0.0 & 78.0 $\pm$ 1.0 \\
\bottomrule
\end{tabular}
\end{center}

\subsection{Decoding temperature}
\label{app:temperature}

All results use greedy decoding.
\Cref{tab:temperature} reports SlimGPT on Qwen3.5-9B at 50\% sparsity under small positive temperatures; a temperature of 0.02 already accounts for most of the loss.

\begin{center}
\small
\captionof{table}{Structured accuracy of SlimGPT on Qwen3.5-9B at 50\% sparsity (BFCL-v3) under greedy decoding and three small positive decoding temperatures.}
\label{tab:temperature}
\begin{tabular}{lcccc}
\toprule
Temperature $\tau$ & 0 (greedy) & 0.02 & 0.05 & 0.10 \\
\midrule
Structured accuracy & 59.1 & 56.9 & 55.3 & 55.3 \\
\bottomrule
\end{tabular}
\end{center}

\subsection{Multi-step healing}
\label{app:multistep}

\Cref{tab:multistep-methods} adds one epoch of healing on 128 HotpotQA trajectories to every 25\%-sparsity model of \cref{tab:HotpotQA}; the dense row is the dense model fine-tuned on the same trajectories.
\Cref{tab:multistep-skim} reports the same step for SKIM on Qwen3.5-2B and 9B together with its BFCL-v3 accuracy, and a variant that instead continues healing on more multi-call trajectories sampled from the xLAM-FC calibration data.
Task-matched trajectory supervision brings SKIM close to the fine-tuned dense model at both scales, while its BFCL-v3 accuracy drops by at most 4.5 points from the single-turn model.

\begin{center}
\small
\captionof{table}{HotpotQA F1 on Qwen3.5-2B at 25\% sparsity before and after one epoch of healing on 128 HotpotQA trajectories; the dense row is the dense model fine-tuned on the same trajectories.}
\label{tab:multistep-methods}
\begin{tabular}{lcc}
\toprule
Method & Single-turn healing & + HotpotQA trajectories \\
\midrule
Dense       & 43.6 & 60.2 \\
\midrule
Wanda-sp    & 14.0 & 26.4 \\
FLAP        & 14.5 & 30.7 \\
SlimGPT     & 19.5 & 32.5 \\
SKIM (ours) & \textbf{20.2} & \textbf{59.7} \\
\bottomrule
\end{tabular}
\end{center}

\begin{center}
\small
\captionof{table}{Multi-step healing of SKIM at 25\% sparsity on Qwen3.5-2B and 9B. BFCL-v3 columns report structured accuracy and HotpotQA columns report F1.}
\label{tab:multistep-skim}
\begin{tabular}{lcccc}
\toprule
& \multicolumn{2}{c}{Qwen3.5-2B} & \multicolumn{2}{c}{Qwen3.5-9B} \\
\cmidrule(lr){2-3} \cmidrule(lr){4-5}
Model & BFCL-v3 & HotpotQA & BFCL-v3 & HotpotQA \\
\midrule
Dense, fine-tuned on HotpotQA trajectories & -- & 60.2 & -- & 75.8 \\
\midrule
SKIM, single-turn healing                  & 80.8 & 20.2 & 85.5 & -- \\
SKIM, + more multi-call trajectories       & 78.3 & 34.5 & 84.0 & 66.2 \\
SKIM, + HotpotQA trajectories              & 78.7 & 59.7 & 81.0 & 73.5 \\
\bottomrule
\end{tabular}
\end{center}

%% file: appendix/prompts.tex
\section{Prompts and Examples}
\label{app:prompts}

We hereby attach below the full prompts (including system prompts) and the output examples for representative success and failure cases on all three datasets: BFCL-v3, API-Bank L1, and HotpotQA. For the two tool-calling datasets (BFCL-v3 and API-Bank L1), the function-call format example and the \texttt{<IMPORTANT>} reminder block in the system prompt are identical across cases and have been replaced with \texttt{[\ldots\ format example and <IMPORTANT> reminders elided \ldots]} in each figure for compactness. The full text of the elided block appears below.

\begin{figure}[h]
\centering
\centering
\begin{tcolorbox}[
  enhanced, sharp corners,
  width=\linewidth,
  colback=promptbg, colframe=promptfr,
  boxrule=0.5pt,
  left=8pt, right=8pt, top=4pt, bottom=4pt,
  before skip=0pt, after skip=0pt,
]
\begin{lstlisting}[
  basicstyle=\ttfamily\scriptsize,
  frame=none,
  breaklines=true,
  columns=fullflexible,
  keepspaces=true,
  xleftmargin=0pt, xrightmargin=0pt,
  aboveskip=0pt, belowskip=0pt,
]
If you choose to call a function ONLY reply in the following format
with NO suffix:

<tool_call>
<function=example_function_name>
<parameter=example_parameter_1>
value_1
</parameter>
<parameter=example_parameter_2>
This is the value for the second parameter
that can span
multiple lines
</parameter>
</function>
</tool_call>

<IMPORTANT>
Reminder:
- Function calls MUST follow the specified format
- Required parameters MUST be specified
- You may provide reasoning in natural language BEFORE the function
  call, but NOT after
- If there is no function call available, answer normally
</IMPORTANT>
\end{lstlisting}
\end{tcolorbox}
\caption{Function-call format example and \texttt{<IMPORTANT>}
reminder block elided from the per-case figures. The block is
token-for-token identical across all tool-calling cases.}
\label{fig:elided-block}
\end{figure}

\begin{figure}[htbp]
  \centering
  \includegraphics[width=0.9\linewidth]{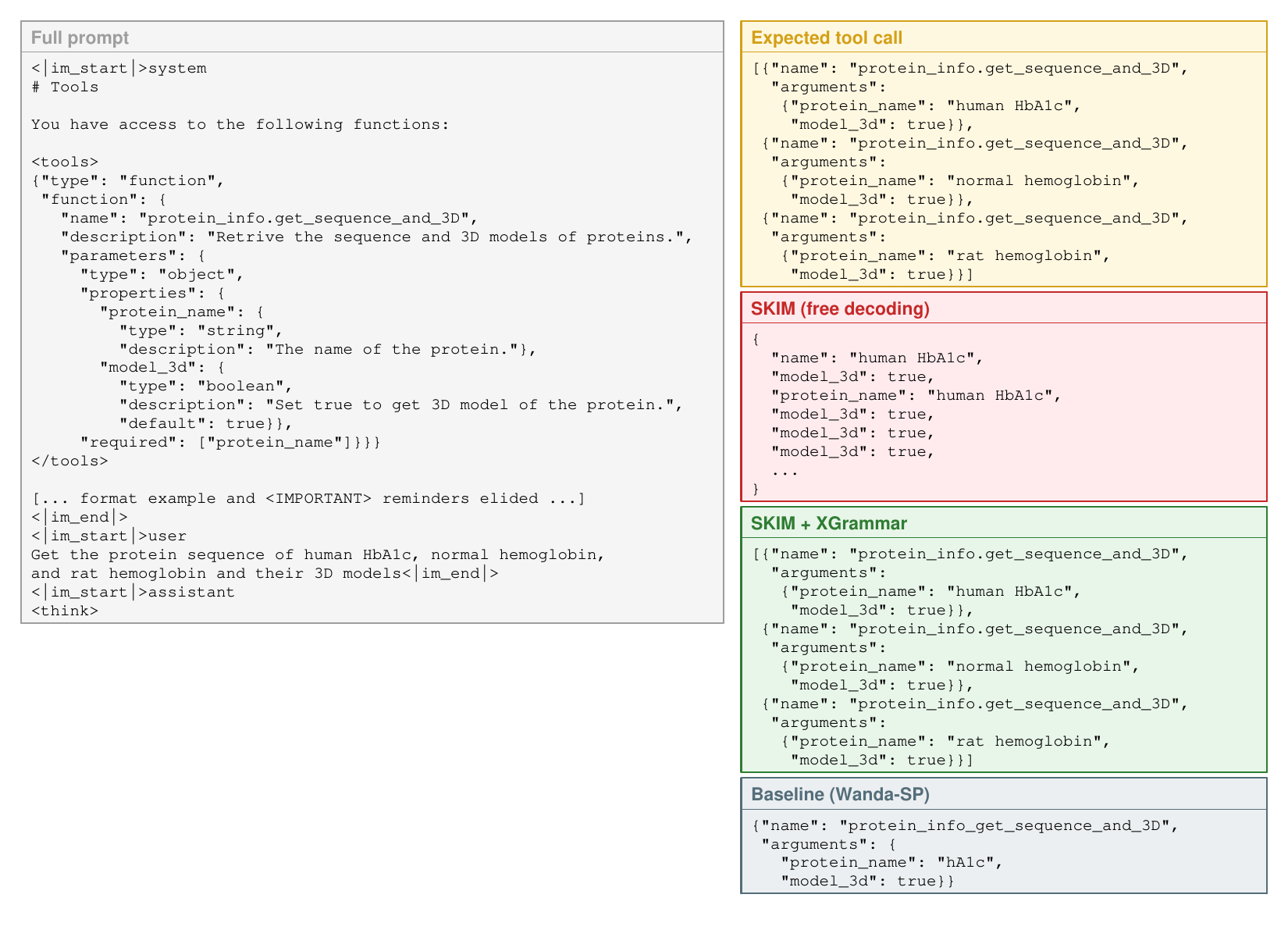}
  \caption{BFCL-v3 success case. SKIM under free decoding collapses into a malformed loop that mixes tool-call fields with argument values; XGrammar restores the exact parallel call set. Wanda-SP at equal density emits one call with a corrupted
  function name and a truncated \texttt{protein\_name}.}
  \label{fig:ex-bfcl-success}
\end{figure}

\begin{figure}[htbp]
  \centering
  \includegraphics[width=0.9\linewidth]{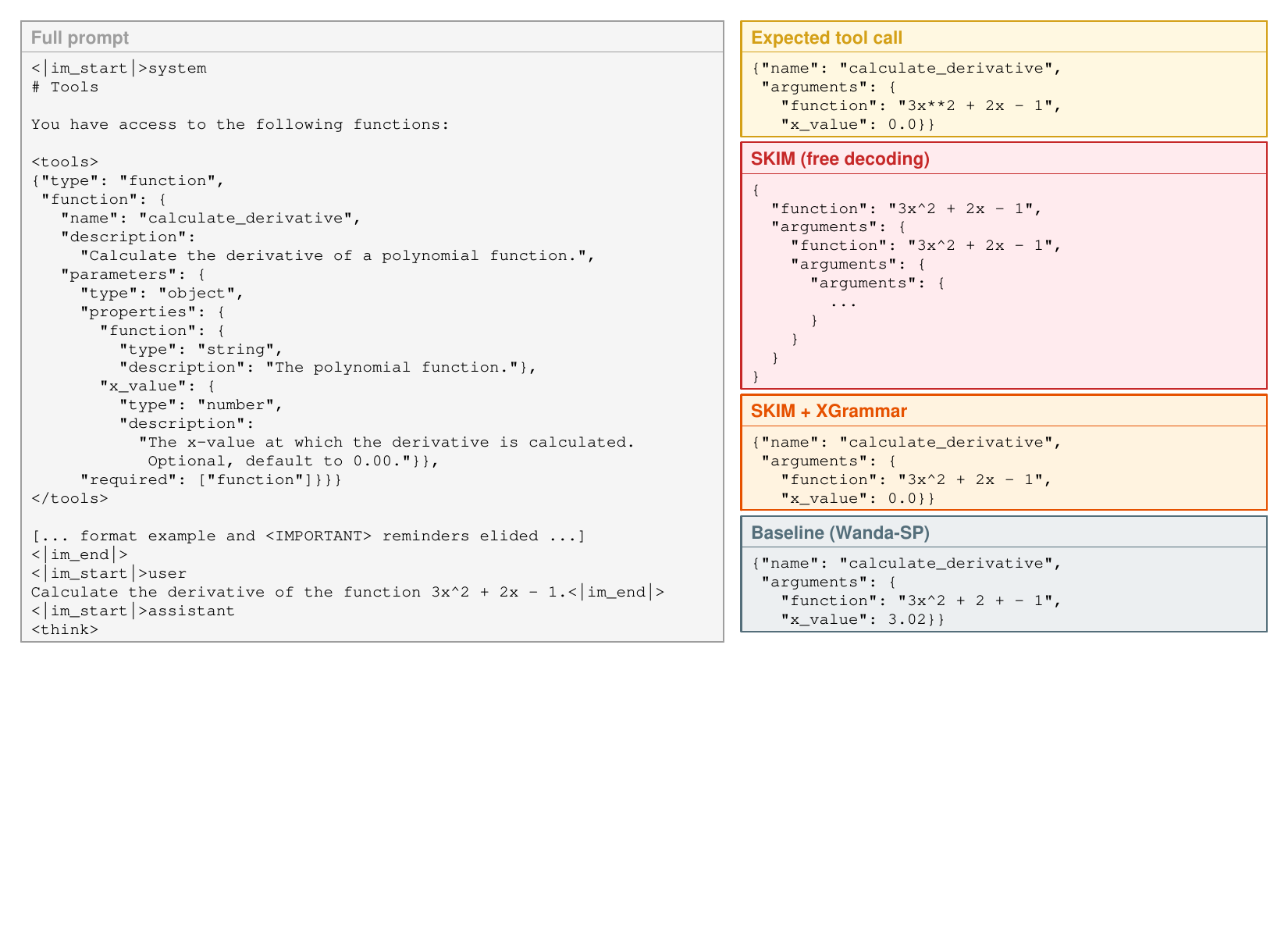}
  \caption{BFCL-v3 failure case. Grammar
  enforcement repairs the structure (well-formed JSON with the right
  function name and argument keys), but the model writes
  \texttt{3x\textasciicircum2} where the gold uses \texttt{3x**2}, so
  the case fails exact match despite format validity. Free decoding
  loops; Wanda-SP corrupts the polynomial expression in the generated call.}
  \label{fig:ex-bfcl-failure}
\end{figure}

\begin{figure}[htbp]
  \centering
  \includegraphics[width=0.9\linewidth]{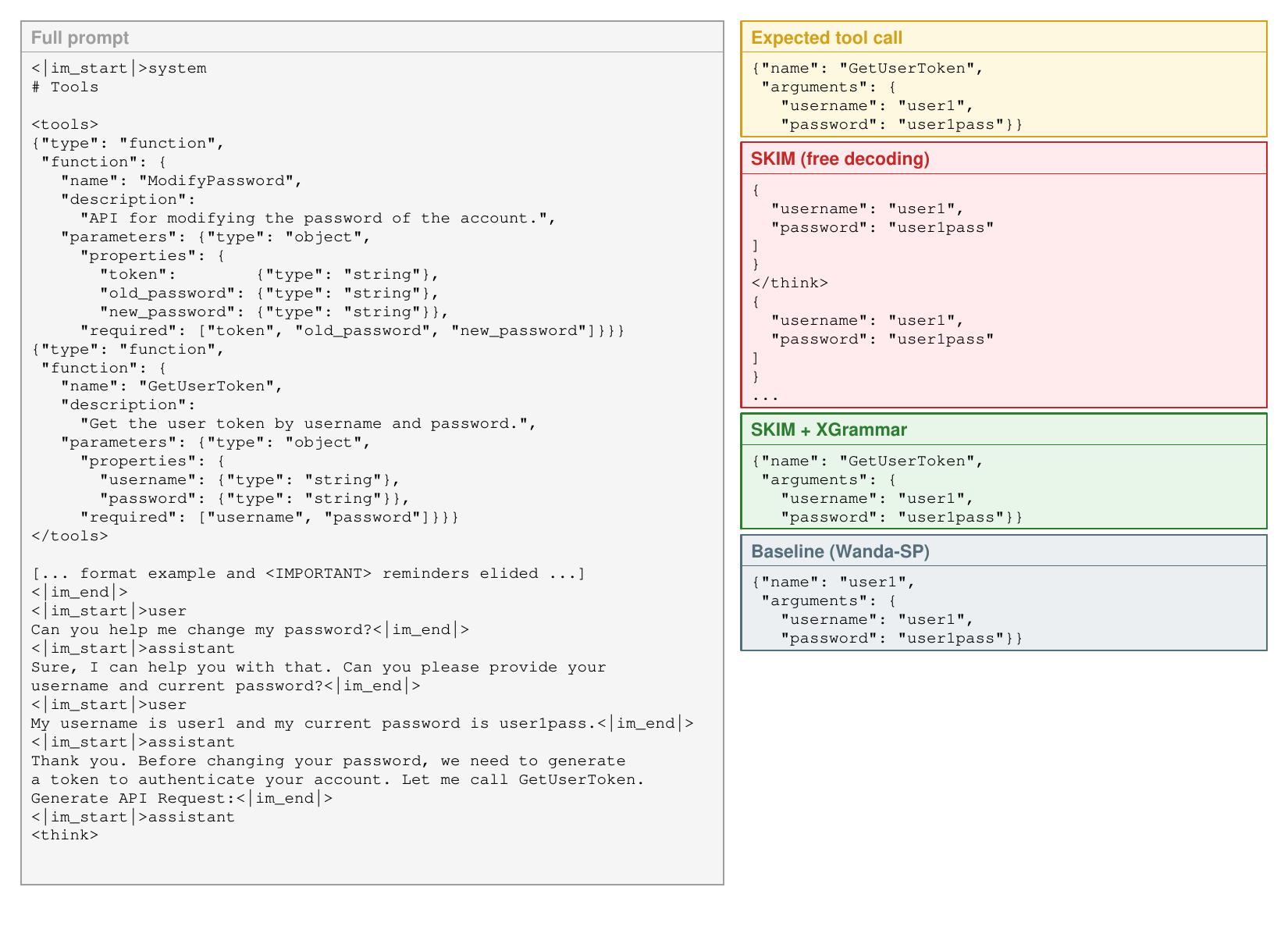}
  \caption{API-Bank L1 success case. The
  multi-turn dialog requires the assistant to authenticate before
  modifying the password; SKIM with XGrammar emits the correct
  \texttt{GetUserToken} call. Free decoding loses the function-name
  slot; Wanda-SP places the username in the \texttt{name} field.}
  \label{fig:ex-apibank-success}
\end{figure}

\begin{figure}[htbp]
  \centering
  \includegraphics[width=0.9\linewidth]{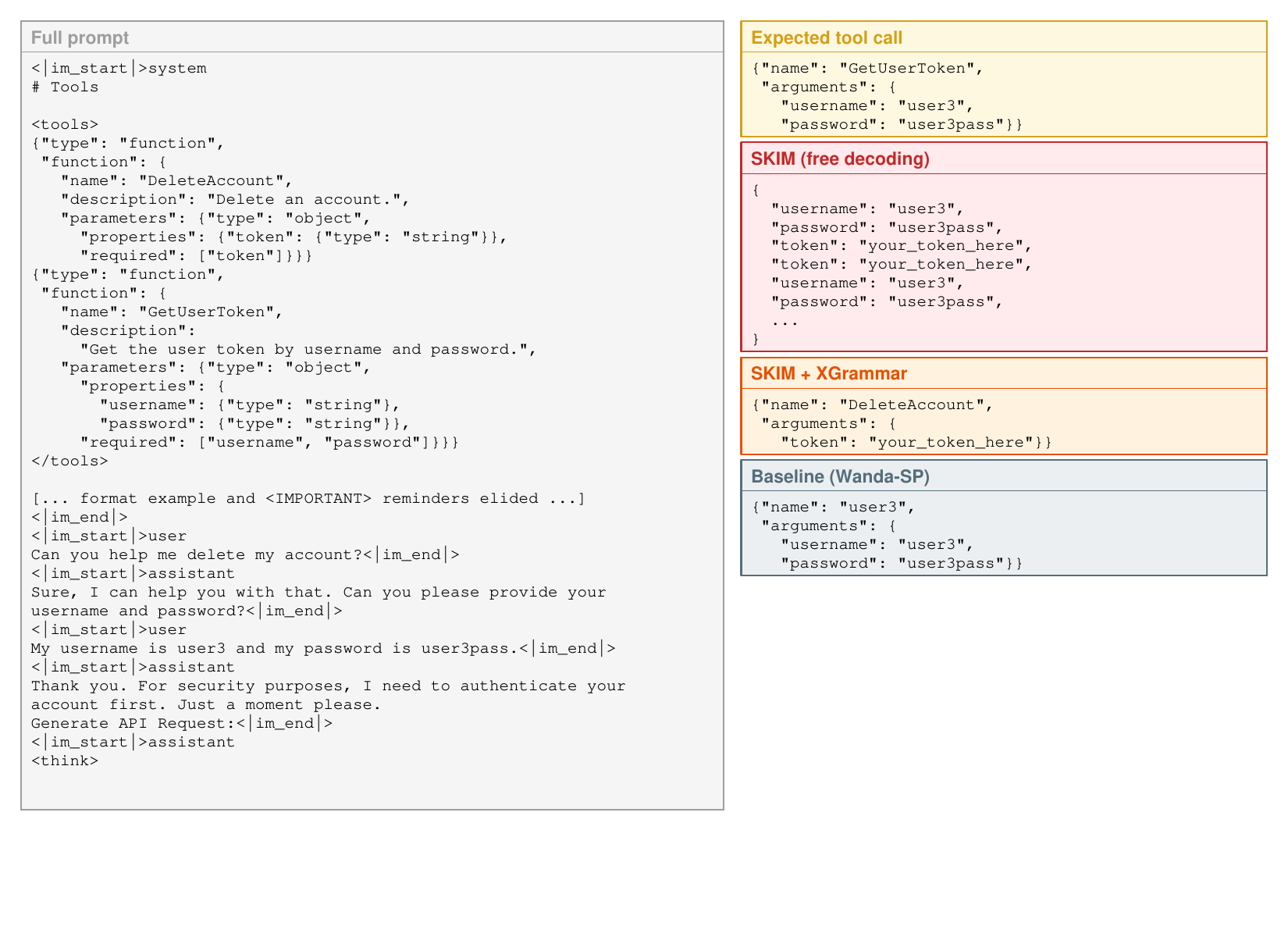}
  \caption{API-Bank L1 failure case. The grammar-constrained decoder produces a perfectly valid call to \texttt{DeleteAccount} with a placeholder \texttt{token}; the user is still in the authentication step and the expected call is \texttt{GetUserToken}. The format is right, the tool is wrong.
  }
  \label{fig:ex-apibank-failure}
\end{figure}

\begin{figure}[htbp]
  \centering
  \includegraphics[width=0.9\linewidth]{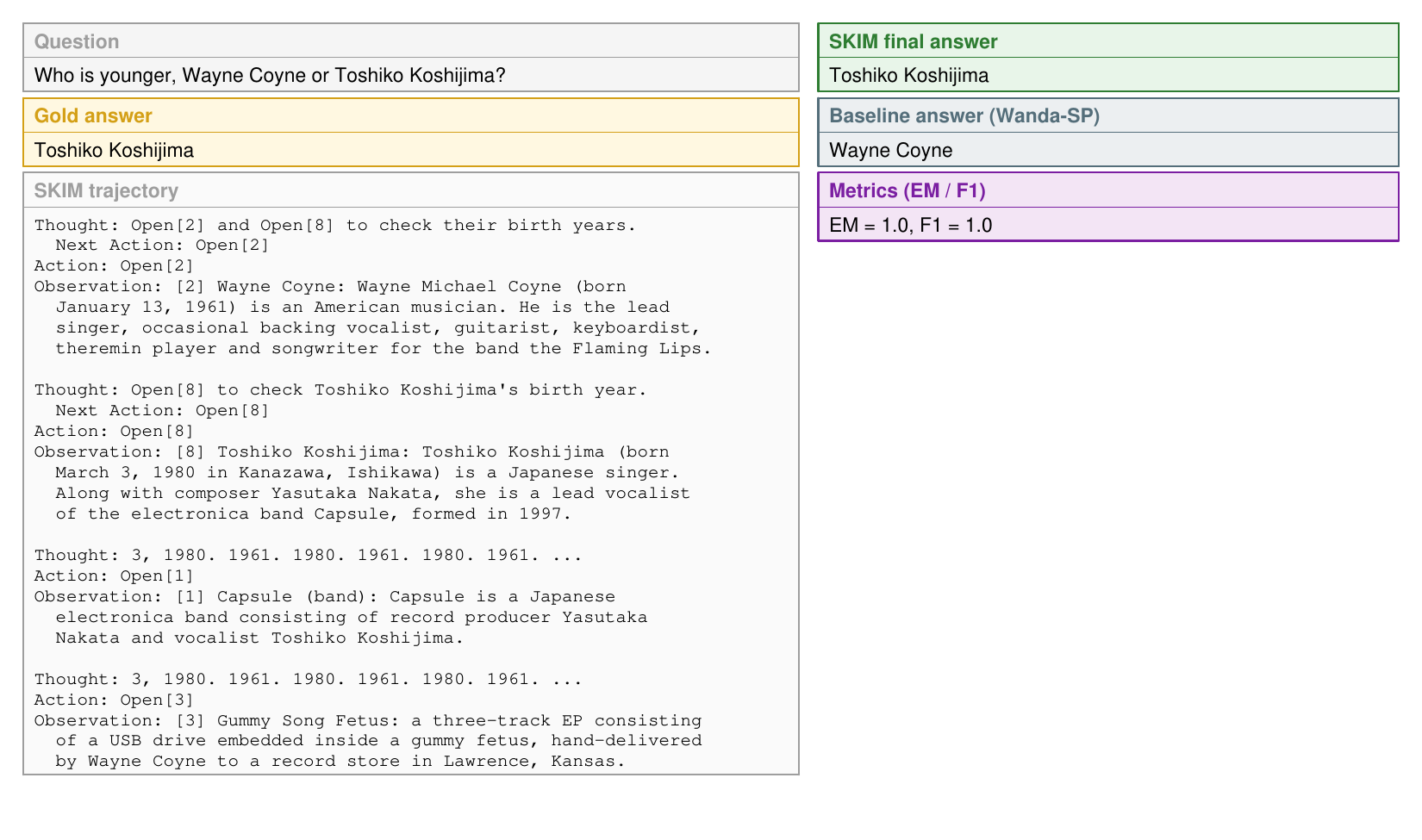}
  \caption{HotpotQA success case.
  The retrieval trajectory is noisy (the model loops through
  \texttt{Open[1]} and \texttt{Open[3]} after already having both
  birth years), but SKIM converges to the correct span;
  EM\,$=$\,1.0, F1\,$=$\,1.0. Wanda-SP returns the wrong person.}
  \label{fig:ex-hotpotqa-success}
\end{figure}

\begin{figure}[htbp]
  \centering
  \includegraphics[width=0.9\linewidth]{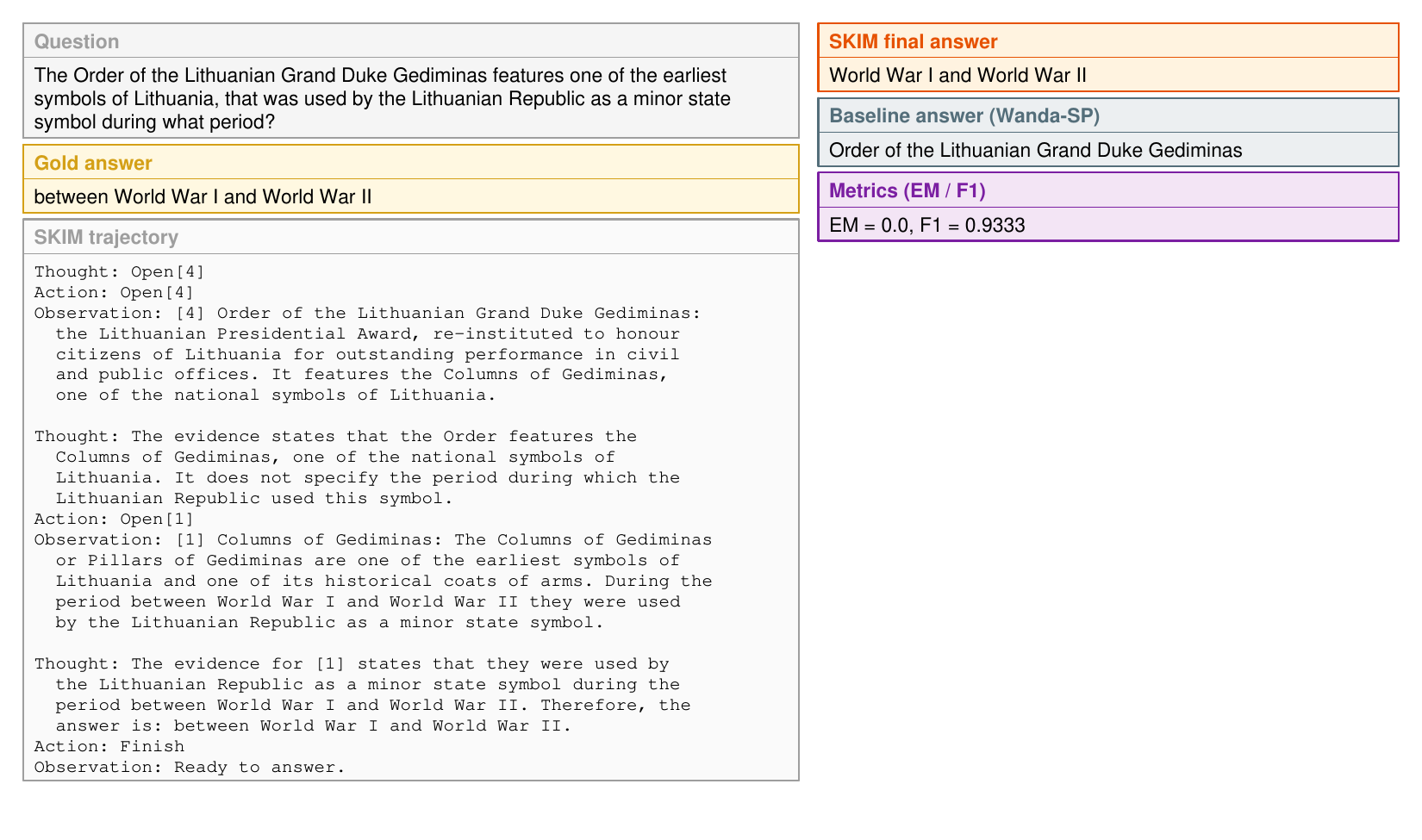}
  \caption{HotpotQA failure case.
  The evidence chain is correct and the model's own reasoning quotes
  the gold span verbatim, but the final answer drops the leading
  preposition ``between''; the answer scores EM\,$=$\,0.0 and F1\,$=$\,0.9333 against the gold span.}
  \label{fig:ex-hotpotqa-failure}
\end{figure}

%% file: appendix/checklist.tex
\section*{NeurIPS Paper Checklist}

\begin{enumerate}

\item {\bf Claims}
    \item[] Question: Do the main claims made in the abstract and introduction accurately reflect the paper's contributions and scope?
    \item[] Answer: \answerYes{}
    \item[] Justification: The abstract and \cref{sec:intro} state our contributions; the structured-decoding gains widening at higher sparsity and the structured-vs-free trade-off claimed there are supported by the experiments in \cref{sec:experiments}.

\item {\bf Limitations}
    \item[] Question: Does the paper discuss the limitations of the work performed by the authors?
    \item[] Answer: \answerYes{}
    \item[] Justification: \cref{sec:discussion} discusses limitations on model scale (up to 27B in the Qwen3.5 series, with full baseline comparisons up to 9B, and Hammer-2.1-3B), distillation data (single-turn function-calling traces only), and pruning-method scope (layer-wise reconstruction, not yet extended to end-to-end pruning or quantization).

\item {\bf Theory assumptions and proofs}
    \item[] Question: For each theoretical result, does the paper provide the full set of assumptions and a complete (and correct) proof?
    \item[] Answer: \answerNA{}
    \item[] Justification: The paper presents an empirical compression method and does not contain formal theorems or proofs requiring an assumption set. However, refer to \cref{sec:skim} for theoretically-motivated proposed method.

    \item {\bf Experimental result reproducibility}
    \item[] Question: Does the paper fully disclose all the information needed to reproduce the main experimental results of the paper to the extent that it affects the main claims and/or conclusions of the paper (regardless of whether the code and data are provided or not)?
    \item[] Answer: \answerYes{}
    \item[] Justification: \cref{sec:experiments} specifies the models, datasets, sparsity levels, baselines, and key hyperparameters; \cref{app:implementation_details} contains the full hyperparameter list and implementation specifics needed for reproduction.

\item {\bf Open access to data and code}
    \item[] Question: Does the paper provide open access to the data and code, with sufficient instructions to faithfully reproduce the main experimental results, as described in supplemental material?
    \item[] Answer: \answerYes{}
    \item[] Justification: The code with configuration files for the main experiments is available on the project page (\url{https://moon4sake.github.io/skim}); the corresponding setup and reproduction instructions are described in \cref{sec:experiments} and \cref{app:implementation_details}.

\item {\bf Experimental setting/details}
    \item[] Question: Does the paper specify all the training and test details (e.g., data splits, hyperparameters, how they were chosen, type of optimizer) necessary to understand the results?
    \item[] Answer: \answerYes{}
    \item[] Justification: \cref{sec:experiments} reports training and evaluation details (data splits, sparsity levels, calibration corpus, distillation hyperparameters, evaluation benchmarks, and decoding settings); full details are deferred to \cref{app:implementation_details}.

\item {\bf Experiment statistical significance}
    \item[] Question: Does the paper report error bars suitably and correctly defined or other appropriate information about the statistical significance of the experiments?
    \item[] Answer: \answerYes{}
    \item[] Justification: \cref{tab:allocation,tab:seeds-methods,tab:seeds-skim} report mean and standard deviation over three seeds for the pruning schedule, the main comparison on Qwen3.5-2B, and SKIM up to 27B. Compute cost prevented running multiple seeds for every (model scale, sparsity level, method) cell; the remaining cells are single runs, and robustness there is supported by trend consistency across four model scales, three sparsity levels, and two backbones in \cref{sec:experiments}.

\item {\bf Experiments compute resources}
    \item[] Question: For each experiment, does the paper provide sufficient information on the computer resources (type of compute workers, memory, time of execution) needed to reproduce the experiments?
    \item[] Answer: \answerYes{}
    \item[] Justification: \cref{app:implementation_details} reports the GPU type and the per-run training cost (e.g., epochs, kd budget, etc).

\item {\bf Code of ethics}
    \item[] Question: Does the research conducted in the paper conform, in every respect, with the NeurIPS Code of Ethics \url{https://neurips.cc/public/EthicsGuidelines}?
    \item[] Answer: \answerYes{}
    \item[] Justification: The work conforms with the NeurIPS Code of Ethics: it uses publicly released models and datasets under their original licenses, does not involve human subjects, and does not raise specific safety, fairness, or privacy concerns beyond those inherent to general LLM compression research.

\item {\bf Broader impacts}
    \item[] Question: Does the paper discuss both potential positive societal impacts and negative societal impacts of the work performed?
    \item[] Answer: \answerNA{}.
    \item[] Justification: Our work is a foundational compression method for existing tool-calling LLMs, and its societal impact is mediated via the underlying models, with no direct application path that introduces additional positive or negative impacts beyond those of the base models being compressed.

\item {\bf Safeguards}
    \item[] Question: Does the paper describe safeguards that have been put in place for responsible release of data or models that have a high risk for misuse (e.g., pre-trained language models, image generators, or scraped datasets)?
    \item[] Answer: \answerNA{}.
    \item[] Justification: The paper does not release pretrained models or scraped datasets that pose a high risk of misuse.

\item {\bf Licenses for existing assets}
    \item[] Question: Are the creators or original owners of assets (e.g., code, data, models), used in the paper, properly credited and are the license and terms of use explicitly mentioned and properly respected?
    \item[] Answer: \answerYes{}
    \item[] Justification: All used models (Qwen3.5 series, Hammer-2.1-3B), datasets (xLAM-FC, BFCL-v3, API-Bank L1, HotpotQA, Spider, GSM8K), and baseline implementations (Wanda, FLAP, SlimGPT) are cited in \cref{sec:experiments}. All their respective licenses are respected.

\item {\bf New assets}
    \item[] Question: Are new assets introduced in the paper well documented and is the documentation provided alongside the assets?
    \item[] Answer: \answerYes{}
    \item[] Justification: The code released on the project page is documented with running bash scripts including training/evaluation matching the experiments in \cref{sec:experiments}.

\newpage
\item {\bf Crowdsourcing and research with human subjects}
    \item[] Question: For crowdsourcing experiments and research with human subjects, does the paper include the full text of instructions given to participants and screenshots, if applicable, as well as details about compensation (if any)?
    \item[] Answer: \answerNA{}.
    \item[] Justification: The work does not involve crowdsourcing or human-subject research.

\item {\bf Institutional review board (IRB) approvals or equivalent for research with human subjects}
    \item[] Question: Does the paper describe potential risks incurred by study participants, whether such risks were disclosed to the subjects, and whether Institutional Review Board (IRB) approvals (or an equivalent approval/review based on the requirements of your country or institution) were obtained?
    \item[] Answer: \answerNA{}.
    \item[] Justification: The work does not involve human-subject research; no IRB approval was required.

\item {\bf Declaration of LLM usage}
    \item[] Question: Does the paper describe the usage of LLMs if it is an important, original, or non-standard component of the core methods in this research? Note that if the LLM is used only for writing, editing, or formatting purposes and does \emph{not} impact the core methodology, scientific rigor, or originality of the research, declaration is not required.
    \item[] Answer: \answerYes{}
    \item[] Justification: GPT-4 is used to label tool-related context tokens in the calibration corpus, as described in \cref{sec:experiments} and \cref{app:implementation_details}. This is a labeling shortcut, not a methodological requirement: a rule-based selection of tool-related spans matches GPT-4 labeling on Qwen3.5-2B and trails it by 1.5--4.6 points on Qwen3.5-9B (\cref{tab:labels}).

\end{enumerate}

%% file: main.bbl
\begin{thebibliography}{47}
\providecommand{\natexlab}[1]{#1}
\providecommand{\url}[1]{\texttt{#1}}
\expandafter\ifx\csname urlstyle\endcsname\relax
  \providecommand{\doi}[1]{doi: #1}\else
  \providecommand{\doi}{doi: \begingroup \urlstyle{rm}\Url}\fi

\bibitem[Agarwal et~al.(2024)Agarwal, Vieillard, Zhou, Stanczyk, Ramos, Geist,
  and Bachem]{agarwal2024policy}
Rishabh Agarwal, Nino Vieillard, Yongchao Zhou, Piotr Stanczyk, Sabela Ramos,
  Matthieu Geist, and Olivier Bachem.
\newblock On-policy distillation of language models: Learning from
  self-generated mistakes.
\newblock In \emph{International Conference on Learning Representations
  (ICLR)}, 2024.

\bibitem[An et~al.(2024)An, Zhao, Yu, Tang, and Wang]{an2024fluctuation}
Yongqi An, Xu~Zhao, Tao Yu, Ming Tang, and Jinqiao Wang.
\newblock Fluctuation-based adaptive structured pruning for large language
  models.
\newblock In \emph{AAAI Conference on Artificial Intelligence (AAAI)}, 2024.

\bibitem[Ashkboos et~al.(2024)Ashkboos, Croci, Nascimento, Hoefler, and
  Hensman]{ashkboos2024slicegpt}
Saleh Ashkboos, Maximilian~L. Croci, Marcelo Gennari~do Nascimento, Torsten
  Hoefler, and James Hensman.
\newblock {SliceGPT}: Compress large language models by deleting rows and
  columns.
\newblock In \emph{International Conference on Learning Representations
  (ICLR)}, 2024.

\bibitem[Beurer-Kellner et~al.(2023)Beurer-Kellner, Fischer, and
  Vechev]{beurer2023prompting}
Luca Beurer-Kellner, Marc Fischer, and Martin Vechev.
\newblock Prompting is programming: A query language for large language models.
\newblock \emph{Proceedings of the ACM on Programming Languages}, 7\penalty0
  (PLDI):\penalty0 1946--1969, 2023.

\bibitem[Cobbe et~al.(2021)Cobbe, Kosaraju, Bavarian, Chen, Jun, Kaiser,
  Plappert, Tworek, Hilton, Nakano, Hesse, and Schulman]{cobbe2021gsm8k}
Karl Cobbe, Vineet Kosaraju, Mohammad Bavarian, Mark Chen, Heewoo Jun, Lukasz
  Kaiser, Matthias Plappert, Jerry Tworek, Jacob Hilton, Reiichiro Nakano,
  Christopher Hesse, and John Schulman.
\newblock Training verifiers to solve math word problems, 2021.
\newblock URL \url{https://arxiv.org/abs/2110.14168}.

\bibitem[Deutsch et~al.(2019)Deutsch, Upadhyay, and Roth]{deutsch2019general}
Daniel Deutsch, Shyam Upadhyay, and Dan Roth.
\newblock A general-purpose algorithm for constrained sequential inference.
\newblock In \emph{Conference on Computational Natural Language Learning
  (CoNLL)}, 2019.

\bibitem[Dong et~al.(2025{\natexlab{a}})Dong, Tang, Liu, Li, Chu, and
  Li]{dong2025can}
Peijie Dong, Zhenheng Tang, Xiang Liu, Lujun Li, Xiaowen Chu, and Bo~Li.
\newblock Can compressed {LLMs} truly act? {An} empirical evaluation of agentic
  capabilities in {LLM} compression.
\newblock In \emph{International Conference on Machine Learning (ICML)},
  2025{\natexlab{a}}.

\bibitem[Dong et~al.(2025{\natexlab{b}})Dong, Ruan, Cai, Lai, Xu, Zhao, and
  Chen]{dong2025xgrammar}
Yixin Dong, Charlie~F. Ruan, Yaxing Cai, Ruihang Lai, Ziyi Xu, Yilong Zhao, and
  Tianqi Chen.
\newblock {XGrammar}: Flexible and efficient structured generation engine for
  large language models.
\newblock In \emph{Machine Learning and Systems (MLSys)}, 2025{\natexlab{b}}.

\bibitem[Frantar and Alistarh(2023)]{frantar2023sparsegpt}
Elias Frantar and Dan Alistarh.
\newblock {SparseGPT}: Massive language models can be accurately pruned in
  one-shot.
\newblock In \emph{International Conference on Machine Learning (ICML)}, 2023.

\bibitem[Frantar et~al.(2023)Frantar, Ashkboos, Hoefler, and
  Alistarh]{frantar2022gptq}
Elias Frantar, Saleh Ashkboos, Torsten Hoefler, and Dan Alistarh.
\newblock {GPTQ}: Accurate post-training quantization for generative
  pre-trained transformers.
\newblock In \emph{International Conference on Learning Representations
  (ICLR)}, 2023.

\bibitem[Geng et~al.(2023)Geng, Josifoski, Peyrard, and West]{geng2023grammar}
Saibo Geng, Martin Josifoski, Maxime Peyrard, and Robert West.
\newblock Grammar-constrained decoding for structured {NLP} tasks without
  finetuning.
\newblock In \emph{Conference on Empirical Methods in Natural Language
  Processing (EMNLP)}, 2023.

\bibitem[Geng et~al.(2025)Geng, Cooper, Moskal, Jenkins, Berman, Ranchin, West,
  Horvitz, and Nori]{geng2025jsonschemabench}
Saibo Geng, Hudson Cooper, Micha{\l} Moskal, Samuel Jenkins, Julian Berman,
  Nathan Ranchin, Robert West, Eric Horvitz, and Harsha Nori.
\newblock {JSONSchemaBench}: A rigorous benchmark of structured outputs for
  language models, 2025.
\newblock URL \url{https://arxiv.org/abs/2501.10868}.

\bibitem[Gu et~al.(2024)Gu, Dong, Wei, and Huang]{gu2024minillm}
Yuxian Gu, Li~Dong, Furu Wei, and Minlie Huang.
\newblock {MiniLLM}: Knowledge distillation of large language models.
\newblock In \emph{International Conference on Learning Representations
  (ICLR)}, 2024.

\bibitem[Han et~al.(2016)Han, Mao, and Dally]{han2015deep_compression}
Song Han, Huizi Mao, and William~J. Dally.
\newblock Deep compression: Compressing deep neural networks with pruning,
  trained quantization and {Huffman} coding.
\newblock In \emph{International Conference on Learning Representations
  (ICLR)}, 2016.

\bibitem[Hao et~al.(2023)Hao, Liu, Wang, and Hu]{hao2023toolkengpt}
Shibo Hao, Tianyang Liu, Zhen Wang, and Zhiting Hu.
\newblock {ToolkenGPT}: Augmenting frozen language models with massive tools
  via tool embeddings.
\newblock In \emph{Conference on Neural Information Processing Systems
  (NeurIPS)}, 2023.

\bibitem[Hinton et~al.(2015)Hinton, Vinyals, and Dean]{hinton2015distilling}
Geoffrey Hinton, Oriol Vinyals, and Jeff Dean.
\newblock Distilling the knowledge in a neural network, 2015.
\newblock URL \url{https://arxiv.org/abs/1503.02531}.

\bibitem[Jiao et~al.(2020)Jiao, Yin, Shang, Jiang, Chen, Li, Wang, and
  Liu]{jiao2020tinybert}
Xiaoqi Jiao, Yichun Yin, Lifeng Shang, Xin Jiang, Xiao Chen, Linlin Li, Fang
  Wang, and Qun Liu.
\newblock {TinyBERT}: Distilling {BERT} for natural language understanding.
\newblock In \emph{Findings of the Association for Computational Linguistics:
  EMNLP}, 2020.

\bibitem[Lew et~al.(2023)Lew, Zhi-Xuan, Grand, and
  Mansinghka]{lew2023sequential}
Alexander~K. Lew, Tan Zhi-Xuan, Gabriel Grand, and Vikash~K. Mansinghka.
\newblock Sequential {Monte Carlo} steering of large language models using
  probabilistic programs, 2023.
\newblock URL \url{https://arxiv.org/abs/2306.03081}.

\bibitem[Li et~al.(2023)Li, Zhao, Yu, Song, Li, Yu, Li, Huang, and
  Li]{li2023api}
Minghao Li, Yingxiu Zhao, Bowen Yu, Feifan Song, Hangyu Li, Haiyang Yu, Zhoujun
  Li, Fei Huang, and Yongbin Li.
\newblock {API-Bank}: A comprehensive benchmark for tool-augmented {LLMs}.
\newblock In \emph{Conference on Empirical Methods in Natural Language
  Processing (EMNLP)}, 2023.

\bibitem[Lin et~al.(2024)Lin, Tang, Tang, Yang, Chen, Wang, Xiao, Dang, Gan,
  and Han]{lin2024awq}
Ji~Lin, Jiaming Tang, Haotian Tang, Shang Yang, Wei-Ming Chen, Wei-Chen Wang,
  Guangxuan Xiao, Xingyu Dang, Chuang Gan, and Song Han.
\newblock {AWQ}: Activation-aware weight quantization for on-device {LLM}
  compression and acceleration.
\newblock In \emph{Machine Learning and Systems (MLSys)}, 2024.

\bibitem[Lin et~al.(2025)Lin, Wen, Peng, Nie, Liao, Wang, Mo, Zhou, Cheng,
  Zhao, Wang, and Zhang]{lin2024hammer}
Qiqiang Lin, Muning Wen, Qiuying Peng, Guanyu Nie, Junwei Liao, Jun Wang,
  Xiaoyun Mo, Jiamu Zhou, Cheng Cheng, Yin Zhao, Jun Wang, and Weinan Zhang.
\newblock {Hammer}: Robust function-calling for on-device language models via
  function masking.
\newblock In \emph{International Conference on Learning Representations
  (ICLR)}, 2025.

\bibitem[Ling et~al.(2024)Ling, Wang, Yan, and Liu]{ling2024slimgpt}
Gui Ling, Ziyang Wang, Yuliang Yan, and Qingwen Liu.
\newblock {SlimGPT}: Layer-wise structured pruning for large language models.
\newblock In \emph{Conference on Neural Information Processing Systems
  (NeurIPS)}, 2024.

\bibitem[Liu et~al.(2024)Liu, Hoang, Zhang, Zhu, Lan, Kokane, Tan, Yao, Liu,
  Feng, Murthy, Yang, Savarese, Niebles, Wang, Heinecke, and
  Xiong]{liu2024apigen}
Zuxin Liu, Thai Hoang, Jianguo Zhang, Ming Zhu, Tian Lan, Shirley Kokane,
  Juntao Tan, Weiran Yao, Zhiwei Liu, Yihao Feng, Rithesh Murthy, Liangwei
  Yang, Silvio Savarese, Juan~Carlos Niebles, Huan Wang, Shelby Heinecke, and
  Caiming Xiong.
\newblock {APIGen}: Automated {PIpeline} for generating verifiable and diverse
  function-calling datasets.
\newblock In \emph{Conference on Neural Information Processing Systems
  (NeurIPS)}, 2024.

\bibitem[Lu et~al.(2022)Lu, Zhang, Chu, Chen, Zhou, Wu, Chen, and
  Yang]{lu2022knowledge}
Chengqiang Lu, Jianwei Zhang, Yunfei Chu, Zhengyu Chen, Jingren Zhou, Fei Wu,
  Haiqing Chen, and Hongxia Yang.
\newblock Knowledge distillation of transformer-based language models
  revisited, 2022.
\newblock URL \url{https://arxiv.org/abs/2206.14366}.

\bibitem[Ma et~al.(2023)Ma, Fang, and Wang]{ma2023llm}
Xinyin Ma, Gongfan Fang, and Xinchao Wang.
\newblock {LLM-Pruner}: On the structural pruning of large language models.
\newblock In \emph{Conference on Neural Information Processing Systems
  (NeurIPS)}, 2023.

\bibitem[Men et~al.(2025)Men, Xu, Zhang, Yuan, Wang, Lin, Lu, Han, and
  Chen]{men2025shortgpt}
Xin Men, Mingyu Xu, Qingyu Zhang, Qianhao Yuan, Bingning Wang, Hongyu Lin,
  Yaojie Lu, Xianpei Han, and Weipeng Chen.
\newblock {ShortGPT}: Layers in large language models are more redundant than
  you expect.
\newblock In \emph{Findings of the Association for Computational Linguistics:
  ACL}, 2025.

\bibitem[Molchanov et~al.(2019)Molchanov, Mallya, Tyree, Frosio, and
  Kautz]{molchanov2019taylor}
Pavlo Molchanov, Arun Mallya, Stephen Tyree, Iuri Frosio, and Jan Kautz.
\newblock Importance estimation for neural network pruning.
\newblock In \emph{IEEE/CVF Conference on Computer Vision and Pattern
  Recognition (CVPR)}, 2019.

\bibitem[Muralidharan et~al.(2024)Muralidharan, Turuvekere~Sreenivas, Joshi,
  Chochowski, Patwary, Shoeybi, Catanzaro, Kautz, and
  Molchanov]{muralidharan2024compact}
Saurav Muralidharan, Sharath Turuvekere~Sreenivas, Raviraj Joshi, Marcin
  Chochowski, Mostofa Patwary, Mohammad Shoeybi, Bryan Catanzaro, Jan Kautz,
  and Pavlo Molchanov.
\newblock Compact language models via pruning and knowledge distillation.
\newblock In \emph{Conference on Neural Information Processing Systems
  (NeurIPS)}, 2024.

\bibitem[{OpenAI}(2023)]{openai2024gpt4technicalreport}
{OpenAI}.
\newblock {GPT-4} technical report, 2023.
\newblock URL \url{https://arxiv.org/abs/2303.08774}.

\bibitem[Patil et~al.(2024)Patil, Zhang, Wang, and Gonzalez]{patil2024gorilla}
Shishir~G. Patil, Tianjun Zhang, Xin Wang, and Joseph~E. Gonzalez.
\newblock {Gorilla}: Large language model connected with massive {APIs}.
\newblock In \emph{Conference on Neural Information Processing Systems
  (NeurIPS)}, 2024.

\bibitem[Patil et~al.(2025)Patil, Mao, Yan, Ji, Suresh, Stoica, and
  Gonzalez]{patil2025berkeley}
Shishir~G. Patil, Huanzhi Mao, Fanjia Yan, Charlie Cheng-Jie Ji, Vishnu Suresh,
  Ion Stoica, and Joseph~E. Gonzalez.
\newblock The {Berkeley} function calling leaderboard ({BFCL}): From tool use
  to agentic evaluation of large language models.
\newblock In \emph{International Conference on Machine Learning (ICML)}, 2025.

\bibitem[Qin et~al.(2024)Qin, Liang, Ye, Zhu, Yan, Lu, Lin, Cong, Tang, Qian,
  Zhao, Hong, Tian, Xie, Zhou, Gerstein, Li, Liu, and Sun]{qin2023toolllm}
Yujia Qin, Shihao Liang, Yining Ye, Kunlun Zhu, Lan Yan, Yaxi Lu, Yankai Lin,
  Xin Cong, Xiangru Tang, Bill Qian, Sihan Zhao, Lauren Hong, Runchu Tian,
  Ruobing Xie, Jie Zhou, Mark Gerstein, Dahai Li, Zhiyuan Liu, and Maosong Sun.
\newblock {ToolLLM}: Facilitating large language models to master 16000+
  real-world {APIs}.
\newblock In \emph{International Conference on Learning Representations
  (ICLR)}, 2024.

\bibitem[{Qwen Team}(2026)]{qwen3.5}
{Qwen Team}.
\newblock {Qwen3.5}: Towards native multimodal agents, 2026.
\newblock URL \url{https://qwen.ai/blog?id=qwen3.5}.

\bibitem[Romero et~al.(2015)Romero, Ballas, Kahou, Chassang, Gatta, and
  Bengio]{romero2014fitnets}
Adriana Romero, Nicolas Ballas, Samira~Ebrahimi Kahou, Antoine Chassang, Carlo
  Gatta, and Yoshua Bengio.
\newblock {FitNets}: Hints for thin deep nets.
\newblock In \emph{International Conference on Learning Representations
  (ICLR)}, 2015.

\bibitem[Sanh et~al.(2019)Sanh, Debut, Chaumond, and Wolf]{sanh2019distilbert}
Victor Sanh, Lysandre Debut, Julien Chaumond, and Thomas Wolf.
\newblock {DistilBERT}, a distilled version of {BERT}: Smaller, faster, cheaper
  and lighter, 2019.
\newblock URL \url{https://arxiv.org/abs/1910.01108}.

\bibitem[Schick et~al.(2023)Schick, Dwivedi-Yu, Dess{\`\i}, Raileanu, Lomeli,
  Hambro, Zettlemoyer, Cancedda, and Scialom]{schick2023toolformer}
Timo Schick, Jane Dwivedi-Yu, Roberto Dess{\`\i}, Roberta Raileanu, Maria
  Lomeli, Eric Hambro, Luke Zettlemoyer, Nicola Cancedda, and Thomas Scialom.
\newblock {Toolformer}: Language models can teach themselves to use tools.
\newblock In \emph{Conference on Neural Information Processing Systems
  (NeurIPS)}, 2023.

\bibitem[Sreenivas et~al.(2024)Sreenivas, Muralidharan, Joshi, Chochowski,
  Mahabaleshwarkar, Shen, Zeng, Chen, Suhara, Diao, Yu, Chen, Ross, Olabiyi,
  Aithal, Kuchaiev, Korzekwa, Molchanov, Patwary, Shoeybi, Kautz, and
  Catanzaro]{sreenivas2024llm}
Sharath~Turuvekere Sreenivas, Saurav Muralidharan, Raviraj Joshi, Marcin
  Chochowski, Ameya~Sunil Mahabaleshwarkar, Gerald Shen, Jiaqi Zeng, Zijia
  Chen, Yoshi Suhara, Shizhe Diao, Chenhan Yu, Wei-Chun Chen, Hayley Ross,
  Oluwatobi Olabiyi, Ashwath Aithal, Oleksii Kuchaiev, Daniel Korzekwa, Pavlo
  Molchanov, Mostofa Patwary, Mohammad Shoeybi, Jan Kautz, and Bryan Catanzaro.
\newblock {LLM} pruning and distillation in practice: The {Minitron} approach,
  2024.
\newblock URL \url{https://arxiv.org/abs/2408.11796}.

\bibitem[Sun et~al.(2024)Sun, Liu, Bair, and Kolter]{sun2023simple}
Mingjie Sun, Zhuang Liu, Anna Bair, and J.~Zico Kolter.
\newblock A simple and effective pruning approach for large language models.
\newblock In \emph{International Conference on Learning Representations
  (ICLR)}, 2024.

\bibitem[Sun et~al.(2019)Sun, Cheng, Gan, and Liu]{sun2019patient}
Siqi Sun, Yu~Cheng, Zhe Gan, and Jingjing Liu.
\newblock Patient knowledge distillation for {BERT} model compression.
\newblock In \emph{Conference on Empirical Methods in Natural Language
  Processing and the International Joint Conference on Natural Language
  Processing (EMNLP-IJCNLP)}, 2019.

\bibitem[Willard and Louf(2023)]{willard2023efficient}
Brandon~T. Willard and R{\'e}mi Louf.
\newblock Efficient guided generation for large language models, 2023.
\newblock URL \url{https://arxiv.org/abs/2307.09702}.

\bibitem[Xia et~al.(2024)Xia, Gao, Zeng, and Chen]{xia2023sheared}
Mengzhou Xia, Tianyu Gao, Zhiyuan Zeng, and Danqi Chen.
\newblock Sheared {LLaMA}: Accelerating language model pre-training via
  structured pruning.
\newblock In \emph{International Conference on Learning Representations
  (ICLR)}, 2024.

\bibitem[Yang et~al.(2025)Yang, Li, Yang, Zhang, Hui, Zheng, Yu, Gao, Huang,
  Lv, et~al.]{yang2025qwen3}
An~Yang, Anfeng Li, Baosong Yang, Beichen Zhang, Binyuan Hui, Bo~Zheng, Bowen
  Yu, Chang Gao, Chengen Huang, Chenxu Lv, et~al.
\newblock {Qwen3} technical report, 2025.
\newblock URL \url{https://arxiv.org/abs/2505.09388}.

\bibitem[Yang et~al.(2018)Yang, Qi, Zhang, Bengio, Cohen, Salakhutdinov, and
  Manning]{yang2018hotpotqa}
Zhilin Yang, Peng Qi, Saizheng Zhang, Yoshua Bengio, William~W. Cohen, Ruslan
  Salakhutdinov, and Christopher~D. Manning.
\newblock {HotpotQA}: A dataset for diverse, explainable multi-hop question
  answering.
\newblock In \emph{Conference on Empirical Methods in Natural Language
  Processing (EMNLP)}, 2018.

\bibitem[Yao et~al.(2023)Yao, Zhao, Yu, Du, Shafran, Narasimhan, and
  Cao]{yao2022react}
Shunyu Yao, Jeffrey Zhao, Dian Yu, Nan Du, Izhak Shafran, Karthik Narasimhan,
  and Yuan Cao.
\newblock {ReAct}: Synergizing reasoning and acting in language models.
\newblock In \emph{International Conference on Learning Representations
  (ICLR)}, 2023.

\bibitem[Yu et~al.(2018)Yu, Zhang, Yang, Yasunaga, Wang, Li, Ma, Li, Yao,
  Roman, Zhang, and Radev]{yu2018spider}
Tao Yu, Rui Zhang, Kai Yang, Michihiro Yasunaga, Dongxu Wang, Zifan Li, James
  Ma, Irene Li, Qingning Yao, Shanelle Roman, Zilin Zhang, and Dragomir Radev.
\newblock {Spider}: A large-scale human-labeled dataset for complex and
  cross-domain semantic parsing and text-to-{SQL} task.
\newblock In \emph{Conference on Empirical Methods in Natural Language
  Processing (EMNLP)}, 2018.

\bibitem[Zhang et~al.(2025)Zhang, Lan, Zhu, Liu, Hoang, Kokane, Yao, Tan, Liu,
  Feng, Niebles, Heinecke, Wang, Savarese, and Xiong]{zhang2025xlam}
Jianguo Zhang, Tian Lan, Ming Zhu, Zuxin Liu, Thai Hoang, Shirley Kokane,
  Weiran Yao, Juntao Tan, Zhiwei Liu, Yihao Feng, Juan~Carlos Niebles, Shelby
  Heinecke, Huan Wang, Silvio Savarese, and Caiming Xiong.
\newblock {xLAM}: A family of large action models to empower {AI} agent
  systems.
\newblock In \emph{Nations of the Americas Chapter of the Association for
  Computational Linguistics (NAACL)}, 2025.

\bibitem[Zheng et~al.(2024)Zheng, Yin, Xie, Sun, Huang, Yu, Cao, Kozyrakis,
  Stoica, Gonzalez, Barrett, and Sheng]{zheng2024sglang}
Lianmin Zheng, Liangsheng Yin, Zhiqiang Xie, Chuyue Sun, Jeff Huang, Cody~Hao
  Yu, Shiyi Cao, Christos Kozyrakis, Ion Stoica, Joseph~E. Gonzalez, Clark
  Barrett, and Ying Sheng.
\newblock {SGLang}: Efficient execution of structured language model programs.
\newblock In \emph{Conference on Neural Information Processing Systems
  (NeurIPS)}, 2024.

\end{thebibliography}
